\documentclass[letterpaper, 10 pt, conference]{ieeeconf}  

\let\labelindent\relax

\usepackage{enumitem}
\usepackage{graphicx, subcaption}
\usepackage{color}
\usepackage{array}
\usepackage{algorithm}
\usepackage{algpseudocode}
\usepackage{amssymb}
\usepackage{soul}
\usepackage[T1]{fontenc} 
\usepackage{amsmath}
\usepackage{hyperref}
\usepackage[colorinlistoftodos]{todonotes}

\usepackage{siunitx}
\usepackage{gensymb}

\renewcommand{\baselinestretch}{1.00}
\newcommand{\colornewparts}[0]{\color{black}}

\newlength\savedwidth
\newcommand{\wcline}[1]{\noalign{\global\savedwidth\arrayrulewidth\global\arrayrulewidth 0.75pt} \cline{#1}
\noalign{\global\arrayrulewidth\savedwidth}}

\makeatletter
 \def\Hline{%
 \noalign{\ifnum0=`}\fi\hrule \@height 1.5pt \futurelet
 \reserved@a\@xhline}
 \makeatother

\IEEEoverridecommandlockouts
\makeatletter
\let\NAT@parse\undefined
\makeatother
\usepackage[numbers]{natbib}

\title{\LARGE \bf
Deep Visual MPC-Policy Learning for Navigation}

\author{Noriaki Hirose, Fei Xia, Roberto Mart\'in-Mart\'in, Amir Sadeghian, Silvio Savarese
\thanks{TOYOTA Central R$\&$D Labs., INC. supported N. Hirose at Stanford University.}
\thanks{The authors are with the Stanford AI Lab, Computer Science Department, Stanford University, 353 Serra Mall, Stanford, CA, USA {\tt\small hirose@mosk.tytlabs.co.jp}
}}

\begin{document}
\maketitle
\thispagestyle{empty}
\pagestyle{empty}

\begin{abstract}
Humans can routinely follow a trajectory defined by a list of images/landmarks. However, traditional robot navigation methods require accurate mapping of the environment, localization, and planning. Moreover, these methods are sensitive to subtle changes in the environment. In this paper, we propose a Deep Visual MPC-policy learning method that can perform visual navigation while avoiding collisions with unseen objects on the navigation path. Our model PoliNet takes in as input a visual trajectory and the image of the robot's current view and outputs velocity commands for a planning horizon of $N$ steps that optimally balance between trajectory following and obstacle avoidance. PoliNet is trained using a strong image predictive model and traversability estimation model in a MPC setup, with minimal human supervision. Different from prior work, PoliNet can be applied to new scenes without retraining. We show experimentally that the robot can follow a visual trajectory when varying start position and in the presence of previously unseen obstacles. We validated our algorithm with tests both in a realistic simulation environment and in the real world. We also show that we can generate visual trajectories in simulation and execute the corresponding path in the real environment. Our approach outperforms classical approaches as well as previous learning-based baselines in success rate of goal reaching, sub-goal coverage rate, and computational load. 
\end{abstract}
%
\section{Introduction}
\label{s:intro}

An autonomously moving agent should be able to reach any location in the environment in a safe and robust manner.
Traditionally, both navigation and obstacle avoidance have been performed using signals from Lidar or depth sensors~\cite{flacco2012depth, fox1997dynamic}. However, these sensors are expensive and prone to failures due to reflective surfaces, extreme illumination or interference~\cite{martin2014deterioration}.
On the other hand, RGB cameras are inexpensive, available on almost every mobile agent, and work in a large variety of environmental and lighting conditions. Further, as shown by many biological systems, visual information suffices to safely navigate the environment.

When the task of moving between two points in the environment is addressed solely based on visual sensor data, it is called \emph{visual navigation}~\cite{zhu2017target}.
In visual navigation, the goal and possibly the trajectory are given in image space.
Previous approaches to visually guide a mobile agent have approached the problem as a control problem based on visual features, leading to \emph{visual servoing} methods~\cite{hutchinson1996tutorial}. However, these methods have a small region of convergence and do not provide safety guarantees against collisions.
{\colornewparts More recent approaches~\cite{zhu2017target,kumar2018visual,pathak2018zero,savinov2018semi} have tackled the navigation problem as a learning task.
Using these methods the agent can navigate different environments. However, reinforcement learning based methods require collecting multiple experiences in the test environment. Moreover, none of these methods explicitly penalize navigating untraversable areas and avoiding collisions and therefore cannot be use for safe real world navigation in changing environments.}

\begin{figure}[t]
  \centering
    \includegraphics[width=0.85\linewidth]{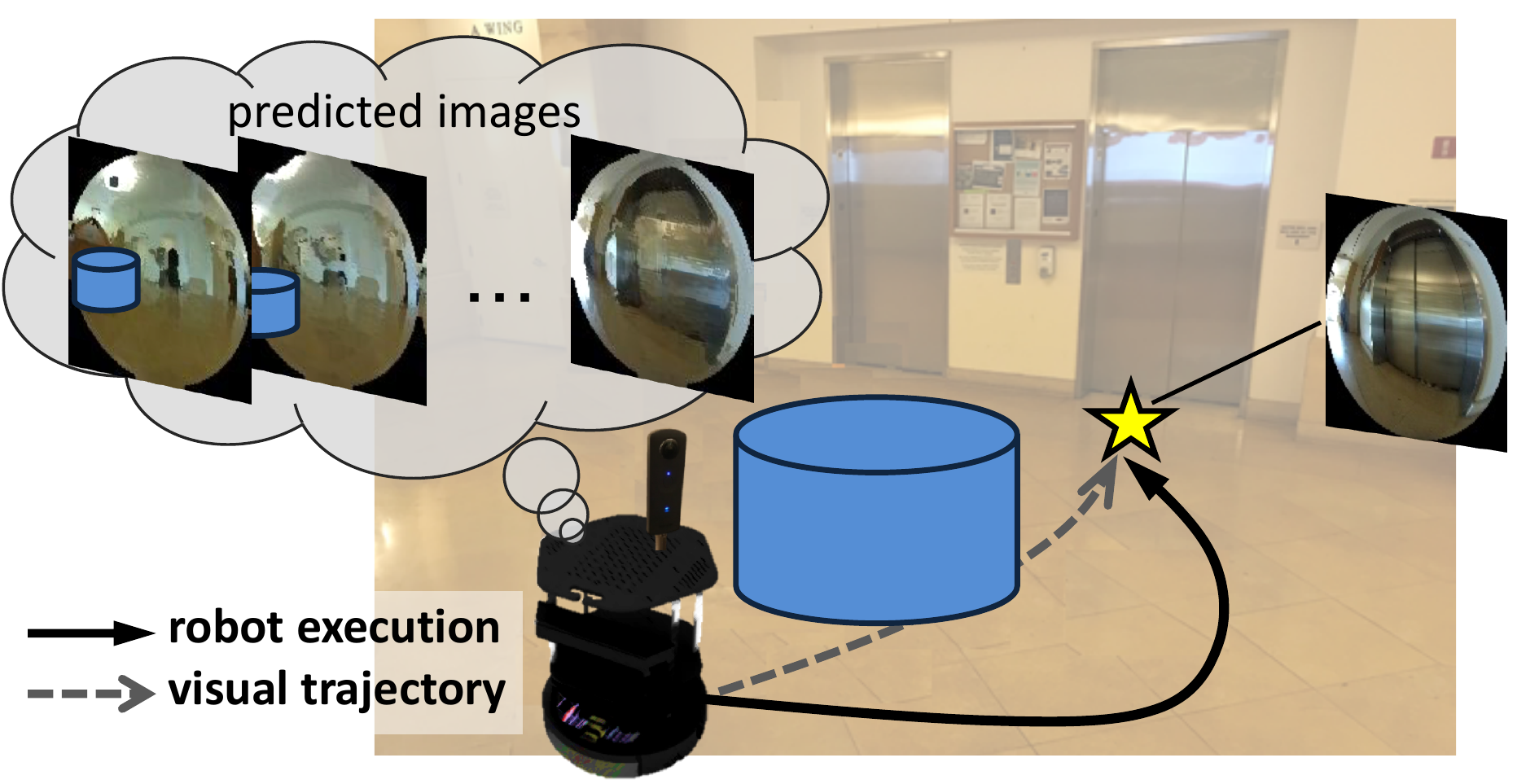}
	\caption{\footnotesize Our robot can navigate in a novel environment following a visual trajectory and deviating to avoid collisions; It uses PoliNet, a novel neural network trained in a MPC setup to generate velocities based on a predictive model and a robust traversability estimator, but that does not require predictions at test time}
	\label{fig:header}
    \vspace{-1.5em}
\end{figure}

In this work, we present a novel navigation system based solely on visual information provided by a \ang{360} camera.
{\colornewparts The main contribution of our work is a novel neural network architecture, PoliNet, that generates the velocity commands necessary for a mobile agent to follow a visual path (a video sequence) while keeping the agent safe from collisions and other risks. 

To learn safe navigation, PoliNet is trained to minimize a model predictive control objective by backpropagating through a differentiable visual dynamics model, VUNet-360, and a differentiable traversability estimation network, GONet, both inspired by our prior work \cite{hirose2019vunet,hirose2018gonet}. PoliNet learns to generate velocities by efficiently optimizing both the path following and traversability estimation objectives.

We evaluate our proposed method and compare to multiple baselines for autonomous navigation on real world and in simulation environments~\cite{xia2018gibson}. We ran a total of 10000 tests in simulation and 110 tests in real world. Our experiments show even in environments that are not seen during training, our agent can reach the goal with a very high success rate, while avoiding various seen/unseen obstacles. Moreover, we show that our model is capable of bridging the gap between simulation and real world; Our agent is able to navigate in real-world by following a trajectory that was generated in a corresponding simulated environment without any retraining.

As part of the development of PoliNet we also propose VUNet-360, a modified version of our prior view synthesis method for mobile robots, VUNet \cite{hirose2019vunet}. VUNet-360 propagates the information between cameras pointing in different directions; in this case, the front and back views of the \ang{360} camera. Finally, we release both the dataset of visual trajectories and the code of the project and hope it can serve as a solid baseline and help ease future research in autonomous navigation \footnote{\href{http://svl.stanford.edu/projects/dvmpc}{http://svl.stanford.edu/projects/dvmpc}}}

\section{Related Work}
\label{rw}

\subsection{Visual Servoing}
\label{rw:vs}
Visual Servoing is the task of controlling an agent's motion so as to minimize the difference between a goal and a current image (or image features)~\cite{hutchinson1996tutorial, espiau1992new, chaumette2006visual, malis19992, chaumette2007visual}.
The most common approach for visual servoing involves defining an Image Jacobian that correlates robot actions to changes in the image space~\cite{hutchinson1996tutorial} and then minimizing the difference between the goal and the current image (using the Image Jacobian to compute a gradient).
There are three main limitations with visual servoing approaches: first, given the \emph{greedy} behavior of the servoing controller, it can get stuck in local minima. Second, direct visual servoing requires the Image Jacobian to be computed, which is costly and requires detailed knowledge about the sensors, agent and/or environment. Third, visual servoing methods only converge well when the goal image can be perfectly recreated through agents actions. In the case of differences in the environment, visual servoing methods can easily break~\cite{1044365}.

The method we present in this paper goes beyond these limitations: we propose to go beyond a pure greedy behavior by using a model predictive control approach. Also, our method does not require expensive Image Jacobian computation but instead learns a neural network that correlates actions and minimization of the image difference. And finally, we demonstrate that our method is not only robust to differences between subgoal and real images, but even robust to the large difference between simulation and real so as to allow sim-to-real transfer.

\subsection{Visual Model Predictive Control}
\label{rw:vmpc}

Model predictive control (MPC)\cite{rawlings2009model} is a multivariate control algorithm that is used
to calculate optimum control moves while satisfying a set of constraints. It can be used when a dynamic model of the process is available. Visual model predictive control studies the problem of model predictive control within a visual servoing scheme\cite{calli2017vision, Tolani:EECS-2018-69, li2016vision, finn2017deep, allibert2008real}. \citet{sauvee2006image} proposed an Image-based Visual Servoing method (IBVS, i.e. directly minimizing the error in image space instead of explicitly computing the position of the robot that would minimize the image error) with nonlinear constraints and a non-linear MPC procedure. In their approach, they measure differences at four pixels on the known objects at the end effector. In contrast, our method uses the differences in the whole image to be more robust against noise and local changes and to capture the complicated scene. 

\citet{finn2017deep} proposed a Visual MPC approach to push objects with a robot manipulator and bring them to a desired configuration defined in image space. Similar to ours, their video predictive model is a neural network. However, to compute the optimal next action they use a sampling-based optimization approach. Compared to \citet{finn2017deep}, our work can achieve longer predictive and control horizon by using a \ang{360} view image and a more efficient predictive model (see Section~\ref{s:expres}). At execution time, instead of a sampling-based optimization method, we use a neural network that directly generates optimal velocity commands; this reduces the high computational load of having to roll out a predictive model multiple times. We also want to point out that due to the partial observable nature of our visual path following problem, it is more challenging than the visual MPC problem of table-top manipulation. 

Another group of solutions, proposed to imitate the control commands of a model-predictive controller (MPC) using neural networks~\cite{lenz2015deepmpc,zhang2016learning,mohamed2019neural,hirose2018mpc}. Differently, we do not train our network PoliNet to imitate an MPC controller but  embed it into an optimization process and train it by backpropagating through differentiable predictive models.

\subsection{Deep Visual Based Navigation}
\label{rw:dvbn}

There has been a surge of creative works in visual-based navigation in the past few years. These came with a diverse set of problem definitions. The common theme of these works is that they don't rely on traditional analytic Image Jacobians nor SLAM-like systems\cite{durrant2006simultaneous, thrun1998probabilistic, thrun2000real, fox1997dynamic} to map and localize and control motion, but rather utilize recent advances in machine learning, perception and reasoning to learn control policies that link images to navigation commands~\cite{gupta2017cognitive, gupta2017unifying, anderson2018vision, brahmbhatt2017deepnav, zhu2017target}. 


A diverse set of tools and methods have been proposed for visual navigation. Learning-based methods including Deep Reinforcement Learning (DRL)~\cite{zhu2017target, kahn2018self, hwangbo2019learning} and Imitation Learning (IL) \cite{codevilla2018end} have been used to obtain navigation policies. Other methods \cite{gupta2017cognitive, paden2016survey} use classical pipelines to perform mapping, localization and planning with separate modules. \citet{chen2018behavioral} proposed a topological representation of the map and a planner to choose behavior. Low level control is behavior-based, trained with behavior cloning. 
{\colornewparts \citet{savinov2018semi} built a dense topological representation from exploration with discrete movements in VizDoom that allows the agent to reach a visual goal.
Among these works, the subset of visual navigation problems most related to our work is the so-called \emph{visual path following}~\cite{kumar2018visual,pathak2018zero}.
Our work compared to these baselines, avoids collisions by explicitly penalizing velocities that brings the robot to areas of low traversability or obstacles. Additionally, we validate our method not only in simulation but also in real world. We show that our agent is able to navigate in real-world by following a trajectory that was generated in a corresponding simulated environment. These experiments show that our method is robust to the large visual differences between the simulation and real world environments. We also show in our evaluation that our method achieves higher success rate than both~\cite{kumar2018visual,pathak2018zero}.}
\begin{figure}[t]
  \centering
    \includegraphics[width=0.9\linewidth]{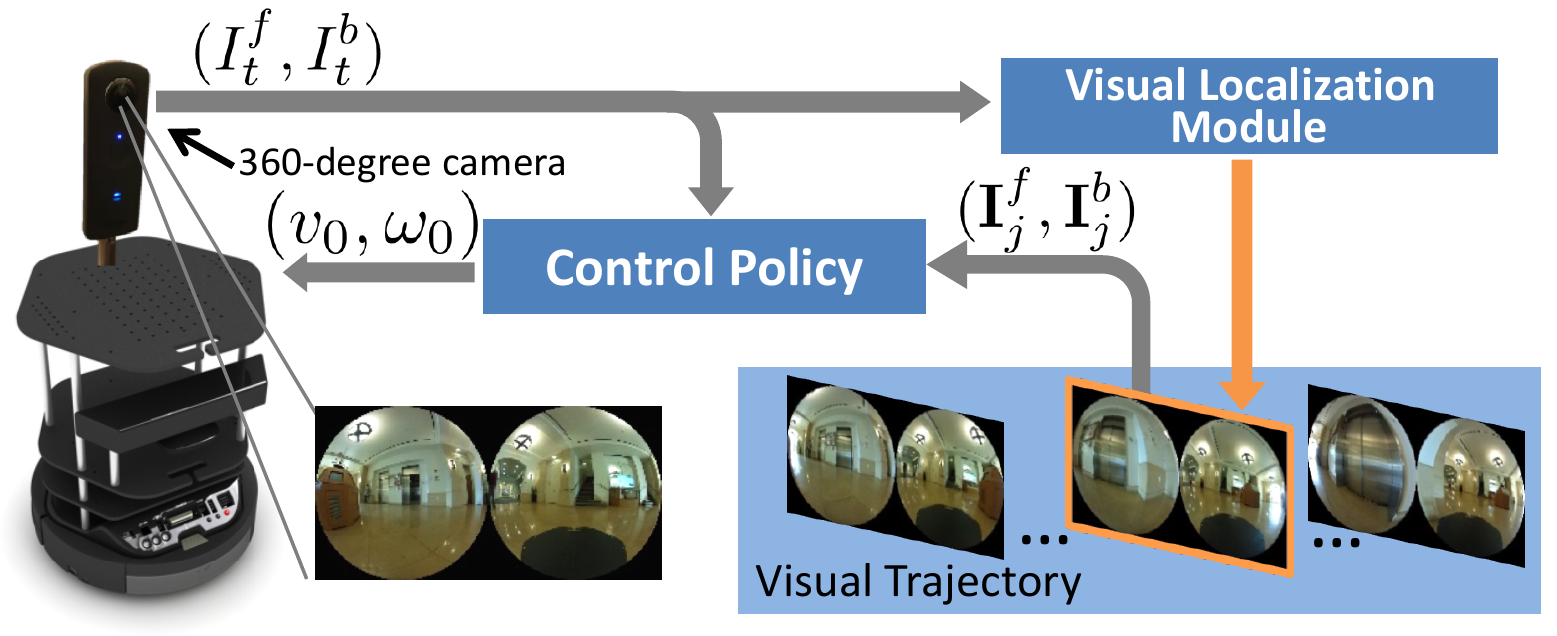}
	\caption{\footnotesize Main process of our method. $I_t^f, I_t^b$ are the front and back side images of \ang{360} camera at time $t$, $\mathbf{I}_j^{f}, \mathbf{I}_j^{b}$ are $j$-th target front and back side images in the visual trajectory, $v_0, \omega_0$ are the linear and the angular velocity for the robot control.}
	\label{fig:flow}
    \vspace{-1.5em}
\end{figure}

\section{Method}
\label{m}

In this section, we introduce the details of our deep visual model predictive navigation approach based on \ang{360} RGB images.

The input to our method is a visual trajectory and a current image, both from a \ang{360} field-of-view RGB camera. The trajectory is defined as consecutive images (i.e. \textbf{subgoals} or waypoints) from a starting location to a target location sampled at a constant time interval. 
We represent the \ang{360} images as two \ang{180} fisheye images (see Fig.~\ref{fig:flow}. Thus, the trajectory can be written as a sequence of $K$ subgoal image pairs, $[(\mathbf{I}_0^{f}, \mathbf{I}_0^{b}),\ldots,(\mathbf{I}_{K-1}^{f}, \mathbf{I}_{K-1}^{b})]$, where the superindex $^f$ indicates \emph{front} and $^b$ indicates \emph{back}. This trajectory can be obtained by teleoperating the real robot or moving a virtual camera in a simulator.

The goal of our control policy is to minimize the difference between the current \ang{360} camera image at time $t$, $(I_t^{f}, I_t^{b})$, and the next subgoal image in the trajectory, $(\mathbf{I}_j^{f}, \mathbf{I}_j^{b})$, while avoiding collisions with obstacles in the environment. These obstacles may be present in the visual trajectory or not, being completely new obstacles. To minimize the image difference, our control policy moves towards a location such that the image from onboard camera looks similar to the next subgoal image. 

A simple heuristic determines if the robot arrived at the current subgoal successfully and switches to the next subgoal. The condition to switch to the next subgoal is the absolute pixel difference between current and subgoal images: $|\mathbf{I}^f_j - I^f_t|+|\mathbf{I}^b_j - I^b_t| < d_{\textrm{\it th}}$, where $j$ is the index of the current subgoal and $d_{\textrm{\it th}}$ is a threshold adapted experimentally. 

The entire process is depicted in Fig.~\ref{fig:flow}. Transitioning between the subgoals our robot can arrive at the target destination without any geometric localization and path planning on a map.

\subsection{Control Policy}
\label{m:polinet}

We propose to control the robot using a model predictive control (MPC) approach in the image domain. However, MPC cannot be solved directly for visual navigation since the optimization problem is non-convex and computationally prohibitive. 
Early stopping the optimization leads to suboptimal solutions (see Section~\ref{s:expres}). We propose instead to learn the MPC-policy with a novel deep neural network we call PoliNet. In the following we first define the MPC controller, which PoliNet is trained to emulate, and then describe PoliNet itself.

PoliNet is trained to minimize a cost function $J$ with two objectives: following the trajectory and moving through traversable (safe) areas. This is in contrast to prior works that only care about following the given trajectory. We propose to achieve these objectives minimizing a linear combination of three losses, the \emph{Image Loss}, the \emph{Traversability Loss} and the \emph{Reference Loss}.
The optimal set of $N$ next velocity commands can be calculated by the minimization of the following cost:
\begin{eqnarray}
\label{eq:cost}
J &= J^{\textrm{\it img}} + \kappa_1 J^{\textrm{\it trav}} + \kappa_2 J^{\textrm{\it ref}}
\end{eqnarray}
\noindent with $\kappa_1$ and $\kappa_2$ constant weights to balance between the objectives.
To compute these losses, we will need to predict future images conditioned on possible robot velocities. We use a variant of our previously presented approach VUNet~\cite{hirose2019vunet} as we explain in Section~\ref{m:pm}. In the following, we will first define the components of the loss function assuming predicted images, followed by the description of our VUNet based predictive model.

\paragraph*{\bf Image Loss}
We define the image loss, $J^{\textrm{\it img}}$, as the mean of absolute pixel difference between the subgoal image $(I_j^{f}, I_j^{b})$ and the sequence of $N$ predicted images $(\hat{I}^f_{t+i}, \hat{I}^b_{t+i})_{i=1 \cdots N}$ as follows:
\begin{eqnarray}
\label{eq:jpixel}
J^{\textrm{\it img}} = \frac{1}{2N\cdot N_{\textrm{\it pix}}} \sum_{i=0}^N w_i(|\mathbf{I}^f_{j} - \hat{I}^f_{t+i}| + |\mathbf{I}^b_{j} - \hat{I}^b_{t+i}|)
\end{eqnarray}
with $N_{\textrm{\it pix}}$ being the number of pixels in the image, $128 \times 128 \times 3$, ($\hat{I}^f_{t+i}, \hat{I}^b_{t+i})_{i=1 \cdots N}$ are predicted images generated by our predictive model (Section~\ref{m:pm}) conditioned on virtual velocities, and $w_i$ weights differently the contributions of consecutive steps for the collision avoidance.

%
\paragraph*{\bf Traversability Loss}
With the traversability loss, $J^{\textrm{\it trav}}$, we aim to penalize areas that constitute a risk for the robot. This risk has to be evaluated from the predicted images.
\citet{hirose2018gonet} presented GONet, a deep neural network-based method that estimates the traversable probability from an RGB image. 
Here we apply GONet to our front predicted images such that we compute the traversability cost based on the traversable probability $\hat{p}^{\textrm{\it trav}}_{t+i}=\textrm{\it GONet}(\hat{I}_{t+i}^{f})$.


{\colornewparts To emphasize the cases with low traversable probability over medium and high probabilities in $J^{\textrm{\it trav}}$, we kernelize the traversable probability as $\hat{p}^{'\textrm{\it trav}} = \textrm{\it Clip}(\kappa^{\textrm{\it trav}}\cdot\hat{p}^{\textrm{\it trav}})$.
The kernelization encourages the optimization to generate commands that avoid areas of traversability smaller than $1/\kappa^{\textrm{\it trav}}$, while not penalizing with cost larger values.}
Based on the kernalized traversable probability, the traversability cost is defined as:
%
\begin{eqnarray}
\label{eq:jgonet}
J^{\textrm{\it trav}} = \frac{1}{N} \sum_{i=0}^N(1 - \hat{p}^{'\textrm{\it trav}}_{t+i})^2
\end{eqnarray}

\paragraph*{\bf Reference Loss}
The image loss and the traversability loss suffice to follow the visual trajectory while avoiding obstacles. However, we observed that the velocities obtained from solving the MPC problem can be sometimes non-smooth and non-realistic since there is no consideration of acceptableness to the real robot in the optimizer.
To generate more realistic velocities we add the reference loss, $J^{\textrm{\it ref}}$, a cost to minimize the difference between the generated velocities $(v_{i}, \omega_{i})_{i=0 \cdots N-1}$ and the real (or simulated) velocities $(v^{\textrm{\it ref}}_{t+i}, \omega^{\textrm{\it ref}}_{t+i})_{i=0 \cdots N-1}$. The reference loss is defines as:
\begin{eqnarray}
\label{eq:cost_all}
J^{\textrm{\it ref}} = \frac{1}{N}\sum_{i=0}^{N-1} (v^{\textrm{\it ref}}_{t+i} - v_i)^2 + \frac{1}{N}\sum_{i=0}^{N-1} (\omega^{\textrm{\it ref}}_{t+i} - \omega_i)^2
\end{eqnarray}
\noindent
This cost is only part of the MPC controller and training process of PoliNet. At inference time PoliNet does not require any geometric information (or the velocities), but only the images defining the trajectory.  

%
%
\begin{figure}[t]
\begin{subfigure}[t]{0.55\hsize}
\includegraphics[width=0.99\hsize]{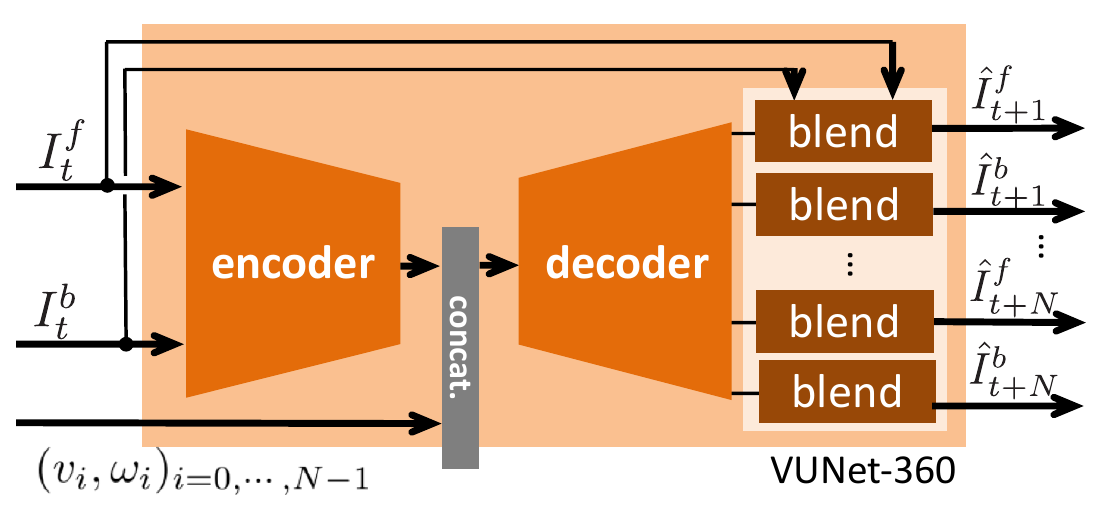}
\caption{\footnotesize VUNet-360 Network}
\end{subfigure}
\begin{subfigure}[t]{0.4\hsize}
\includegraphics[width=0.99\hsize]{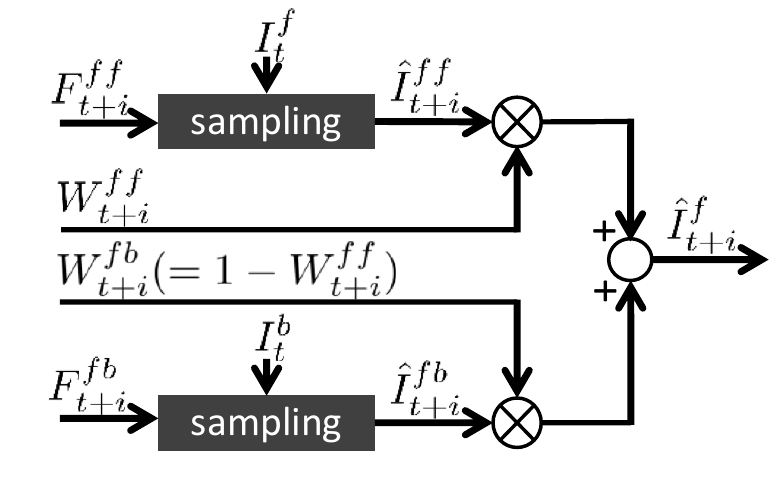} 
\caption{\footnotesize Blending module in detail}
\end{subfigure}
\caption{\footnotesize Network structure of VUNet-360 comprising a encoder-decoder with concatenation of virtual velocities in the latent space for conditioning and a blending mechanism (left); A novel blending module (details on the right) exploits the corresponding part of the front and back images to generate full front and back virtual 360 images}
\label{fig:vunet}
\vspace{-15pt}
\end{figure}

%

\subsection{Predictive Model, VUNet-360}
\label{m:pm}

The previously described loss function requires a forward model that generates images conditioned on virtual velocities. Prior work VUNet~\cite{hirose2019vunet} proposed a view synthesize approach to predict future views for a mobile robot given a current image and robot's possible future velocities. 
{\colornewparts However, we cannot use VUNet as forward model to train PoliNet within the previously defined MPC because: 1) the original VUNet uses and predicts only a front camera view, and 2) the multiple step prediction process in VUNet is sequential, which not only leads to heavy computation but also to lower quality predictions. We address these problems we propose \emph{VUNet-360}, a modified version of VUNet that uses as input one \ang{360} image (in the form of two \ang{180} images) and $N$ virtual velocities, and generates in parallel $N$ predicted future images.}
%

The network structure of VUNet-360 is depicted in Fig.~\ref{fig:vunet}[a]. Similar to the original VUNet, our version uses an encoder-decoder architecture with robot velocities concatenated to the latent vector. One novelity of VUNet-360 the computation is now in parallel for all $N$ images. We also introduced the blending module that generates virtual front and back images by fusing information from the input front and back images.

Fig.~\ref{fig:vunet}[b] shows $i$-th blending module for the prediction of $\hat{I}_{t+i}^f$. Similar to \cite{zhou2016view}, we use bilinear sampling to map pixels from input images to predicted images.

The blending module receive 2 flows, $F_{t+i}^{ff}, F_{t+i}^{fb}$ and 2 visibility masks $W_{t+i}^{ff}, W_{t+i}^{fb}$ from the decoder of VUNet-360. This module blends the sampled front and back images $\hat{I}_{t+i}^{ff}, \hat{I}_{t+i}^{fb}$ by $F_{t+i}^{ff}, F_{t+i}^{fb}$ with the masks $W_{t+i}^{ff}, W_{t+i}^{fb}$ to use both front and back image pixels to predict $\hat{I}_{t+i}^f$.


To train VUNet-360 we input a real image $(I^f_{t}, I^b_{t})$ and a sequence of real velocities $(v_{i}, \omega_{i})_{i=0 \cdots N-1} = (v_{t+i}, \omega_{t+i})_{i=0 \cdots N-1}$ collected during robot teleoperation (see Sec.~\ref{s:es}) and we minimize the following cost function:
\begin{eqnarray}
J^{\textrm{\it VUNet}} = \frac{1}{2N\cdot N_{\textrm{\it pix}}}\sum_{i=1}^N (|I^f_{t+i} - \hat{I}^f_{t+i}| + |I^b_{t+i} - \hat{I}^b_{t+i}|)
\end{eqnarray}
\noindent
where $(I^f_{t+i}, I^b_{t+i})_{i=1 \cdots N}$ are the ground truth future images and $(\hat{I}^f_{t+i}, \hat{I}^b_{t+i})_{i=1 \cdots N}$ are the VUNet predictions.

\subsection{Neural Model Control Policy, PoliNet}
\label{m:cp}

\begin{figure}[t]
  \centering
    \includegraphics[width=0.99\hsize]{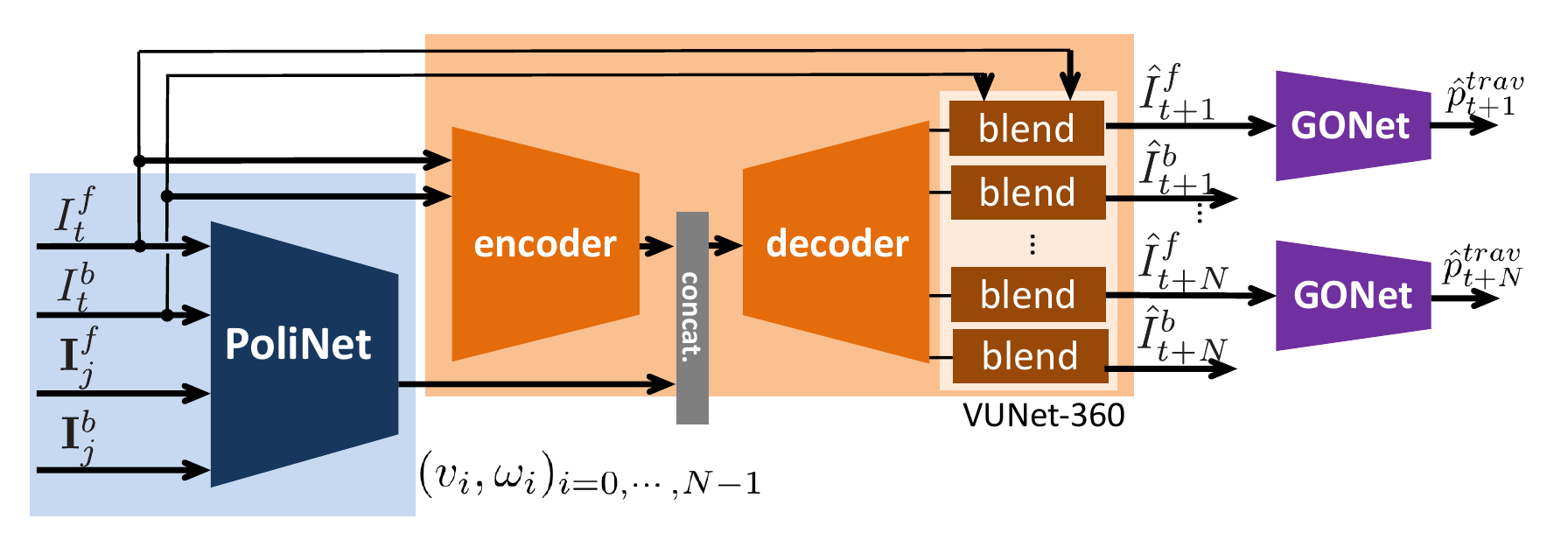}
	\caption{\footnotesize Forward calculation in the training process of PoliNet. Generated images based on PoliNet velocities, and traversability values are used to compute the loss. At test time only the blue shadowed area is used}
	\label{fig:training}
    \vspace{-1.5em}
\end{figure}

The optimization problem of the model predictive controller with the predictive model described above cannot be solved online within the required inference time due to the complexity (non-convexity) of the minimization of the cost function.
We propose to train a novel neural network, PoliNet in the dashed black rectangle of Fig.\ref{fig:training}, and only calculate PoliNet online to generate the velocities instead of the optimization problem.
The network structure of PoliNet is simply constructed with 8 convolutional layers to allow fast online computation on the onboard computer of our mobile robot.
In the last layer, we have tanh($\cdot$) to limit the linear velocity within $\pm v_{max}$ and the angular velocity within $\pm \omega_{max}$.


Similar to the original MPC, the input to PoliNet is the current image, $(I_t^{f}, I_t^{b})$, and the subgoal image, $(\mathbf{I}^f_j, \mathbf{I}^b_j)$, and the output is a sequence of $N$ robot velocities, $(v_{t+i}, \omega_{t+i})_{i=0 \cdots N-1}$, that move the robot towards the subgoal in image space while keeping it away from non-traversable areas.
By forward calculation of PoliNet, VUNet-360, and GONet as shown in Fig.\ref{fig:training}, we can calculate the same cost function $J$ as MPC-policy to update PoliNet.
Note that VUNet-360 and GONet are not updated during the training process of PoliNet. VUNet-360 and GONet are only used to calculate the gradient to update PoliNet.
To train PoliNet, we need $(I_t^{f}, I_t^{b})$ as current image,  $(\mathbf{I}_j^{f}, \mathbf{I}_j^{b})$ as the subgoal image, and the tele-operator's velocities $(v_{i}, \omega_{i})_{i=0 \cdots N-1}=(v_{t+i}, \omega_{t+i})_{i=0 \cdots N-1}$ for $J_{ref}$.
We randomly choose the future image from {\colornewparts the dataset} as the target image $(\mathbf{I}_j^{f}, \mathbf{I}_j^{b}) = (I_{t+k}^f, I_{t+k}^b)$.
Here, $k$ is the random number within $N_r$.

\section{Experimental Setup}
\label{s:es}

We evaluate our deep visual navigation approach based on our learned visual MPC method as the navigation system for a Turtlebot 2 with a Ricoh THETA S 360 camera on top. 
We will conduct experiments both in real world and in simulation using this robot platform.

To train our VUNet-360 and PoliNet networks we collect new data both in simulation and in real world. In real world we teleoperate the robot and collect 10.30 hours of \ang{360} RGB images and velocity commands in twelve buildings at the Stanford University campus. We separate the data from the different buildings into data from eight buildings for training, from two buildings for validation, and from two other buildings for testing.

In simulation we use the GibsonEnv~\cite{xia2018gibson} simulator with models from Stanford 2D-3D-S~\cite{armeni2017joint} and Matterport3D~\cite{chang2017matterport3d} datasets. The models from Stanford 2D-3D-S are office buildings at Stanford reconstructed using a 3D scanner with texture. This means, for these buildings we have corresponding environments in simulation and real world. Differently, Matterport3D mainly consists of residential buildings. Training on both datasets gives us better generalization performance to different types of environments. 

In the simulator, we use a virtual \ang{360} camera with intrinsic parameters matching the result of a calibration of the real Ricoh THETA S camera. We also teleoperate the virtual robot in 36 different simulated buildings for 3.59 hours and split the data into 2.79 hours of training data from 26 buildings, 0.32 hours of data as validation set from another 5 buildings, and 0.48 hours of data as test set from another 5 buildings.

To train VUNet-360 and PoliNet, we balance equally simulator's and real data.
We train iteratively all networks using Adam optimizer with a learning rate of 0.0001 on a Nvidia GeForce RTX 2080 Ti GPU. All collected images are resized into 128$\times$128 before feeding into the network.

\begin{figure*}[t]
  \centering
    
    \begin{tabular}{ccc}
      \includegraphics[width=0.37\hsize]{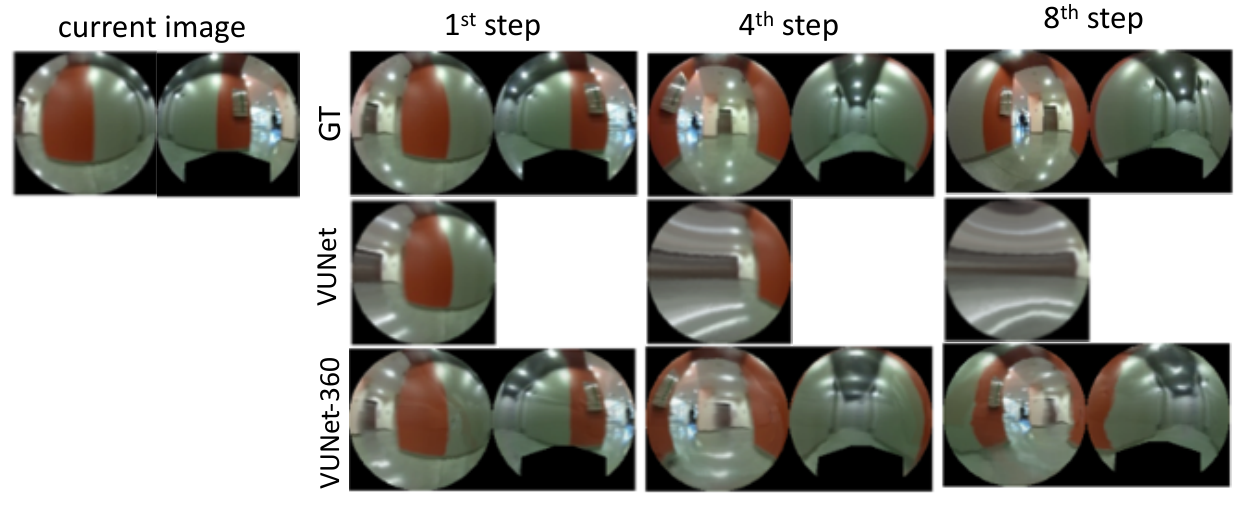} & \includegraphics[width=0.37\hsize]{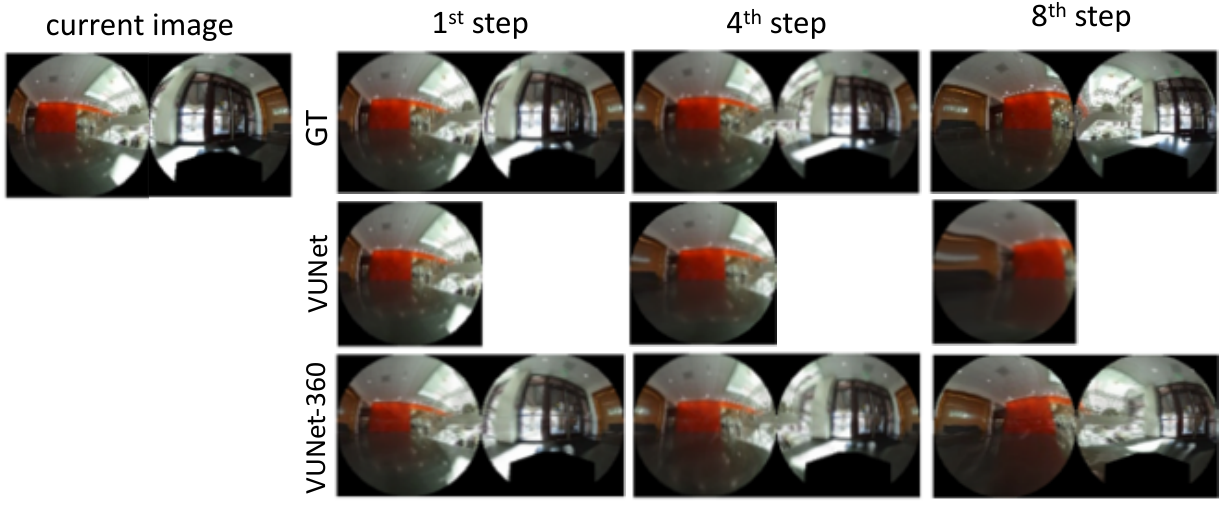} & \includegraphics[width=0.18\hsize]{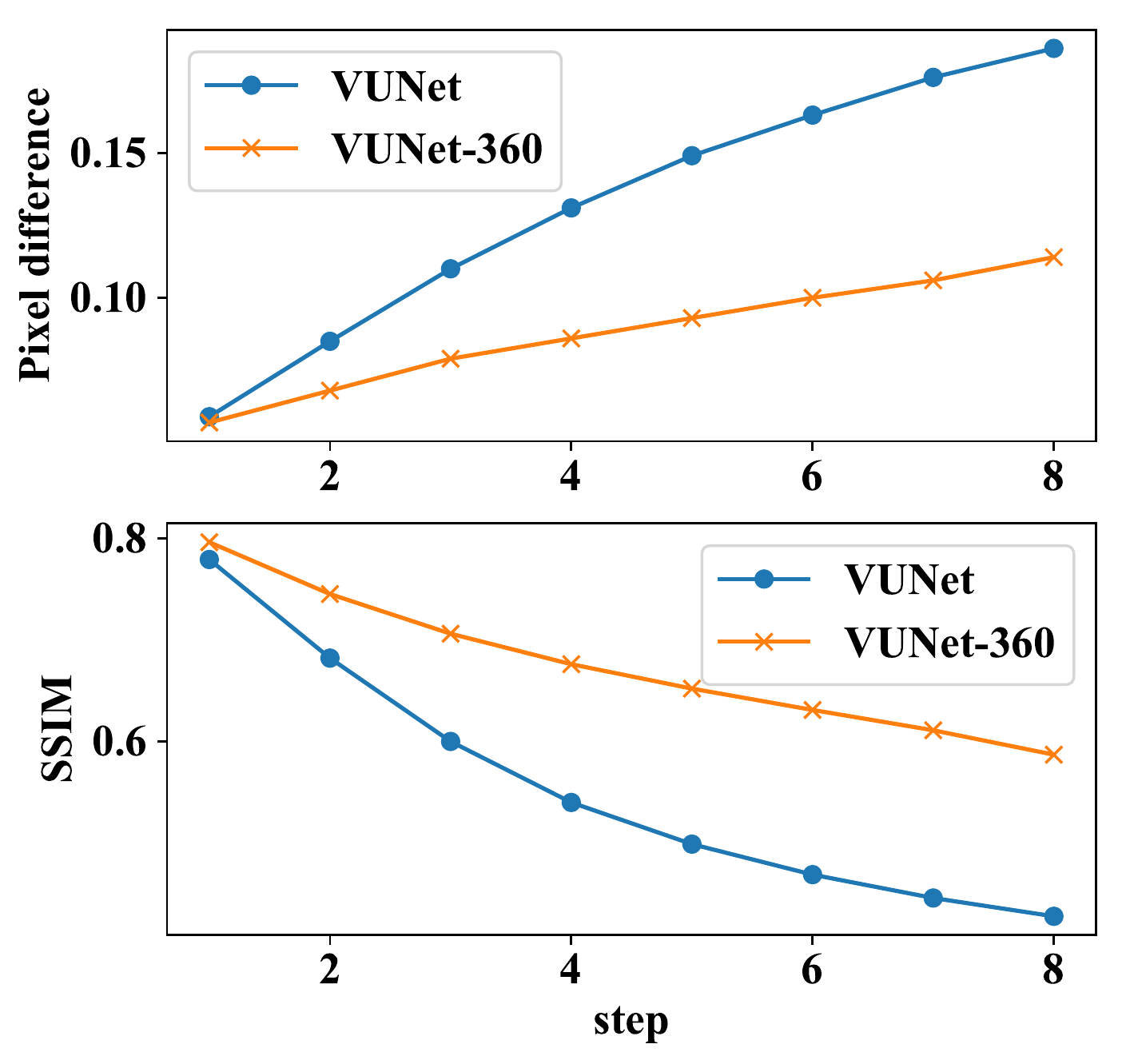} \\
      {\small $\left[ \mbox{a} \right]$} & {\small $\left[ \mbox{b} \right]$ } & {\small $\left[ \mbox{c} \right]$ } \\
    \end{tabular}

	\caption{\footnotesize Evaluation of our proposed predictive model, VUNet-360, and comparison to the original model VUNet~\cite{hirose2018gonet++}. [a][b] show ground truth images (top) and predicted images from VUNet (middle) and VUNet-360 (bottom) while rotating in place ([a]) or moving forward and slightly rotating ([b]). [c] shows SSIM and pixel difference for VUNet and VUNet-360. VUNet-360 improves the predictive capabilities through the right exploitation of information from the back image.}
	\label{fig:img_vunet}
    \vspace{-1.5em}
\end{figure*}

\paragraph*{Parameters}

We set $N=8$ steps as horizon and \SI{0.333}{\second} (\SI{3}{\hertz}) as inference time. This inference time allows to use our method in real time navigation. The values correspond to a prediction horizon of \SI{2.667}{\second} into the future. These values are a good balance between long horizon that is generally better in MPC setups, and short predictions that are more reliable with a predictive model as VUNet-360 (see Fig. \ref{fig:img_vunet}).

$v_{max}$ and $\omega_{max}$ are given as \SI{0.5}{\meter/\second} and \SI{1.0}{\radian/\second}. Hence, the maximum range of the prediction can be $\pm$ \SI{1.333}{\meter} and $\pm$ \SI{2.667}{\radian} from the current robot pose, which are large enough to allow the robot  avoiding obstacles.

{\colornewparts In $J^{\textrm{\it img}}$, $w_i = 1.0$ for $i\neq N$ and $w_N = 5.0$ to allow deviations from the original visual trajectory to avoid collisions while encouraging final convergence.}
$\kappa^{\textrm{\it trav}}$ for $J^{\textrm{\it trav}}$ is set to 1.1 and {\colornewparts $d_{\textrm{\it th}}$ is defined as $k_{\textrm{\it th}} (|\mathbf{I}^f_j - I^f_t|+|\mathbf{I}^b_j - I^b_t|)$ when switching the subgoal image. $k_{\textrm{\it th}}$ is experimentally set to 0.7.}
The weight of the different terms in the cost function (Eq.~\ref{eq:cost}) are $\kappa_1=0.5$ for the traversability loss and $\kappa_2=0.1$ for the reference loss. The optimal $\kappa_1$ is found through ablation studies (see Table~\ref{tab:ablation}). Setting $\kappa_2=0.1$, we limit the contribution of the reference loss to the overall cost because our goal with this additional term is not to learn to imitate the teleoperator's exact velocity but to regularize and obtain smooth velocities.
In addition, we set $N_r=12$ to randomly choose the subgoal image for the training of PoliNet.

\section{Experiments}
\label{s:expres}

We conduct three sets of experiments to validate our method in both simulation and real world. We first evaluate the predictive module: VUNet-360. Then we evaluate the performance of PoliNet by comparing it against a set of baselines, both as a MPC-learned method and as a core component of our proposed deep visual navigation approach. Finally, we perform ablation studies to understand the importance of the traversability loss computed by GONet in our loss design.

\subsection{Evaluation of VUNet-360}

\paragraph*{Quantitative Analysis}
First we evaluate the quality of the predictions from our trained VUNet-360 on the images of the test set, and compare to the original method, VUNet. We use two metrics: the pixel difference (lower is better) and SSIM (higher is better). As is shown in  Fig.~\ref{fig:img_vunet}~[c], VUNet-360 clearly improves over the original method in all stages of the prediction.

We also compare the computation efficiency of VUNet and VUNet-360, since this is relevant for efficient training of PoliNet. VUNet-360 has smaller memory footprint (901MB vs. 1287MB) and higher frequency (\SI{19.28}{\hertz} vs. \SI{5.03}{\hertz}) than VUNet. This efficiency gain is because VUNet-360 predicts multiple images in parallel with a single forward pass of the encoder-decoder network.

\paragraph*{Qualitative Analysis}

Fig.~\ref{fig:img_vunet}[a,b] shows predicted images from VUNet and VUNet-360 for two representative scenarios.
In the images for each scenario, on the left side we show the current real front and back images that compose the 360\degree image. From second to fourth column, we show the ground truth image, VUNet prediction and VUNet-360 prediction for 1, 4 and 8 steps. In the first scenario, the robot is turning in place. In the second scenario, the robot is moving forward while slightly turning left. We observe that VUNet-360 more accurately predicts future images and better handles occlusions since it is able to exploit information from the back camera.

\begin{figure*}[t]
  \begin{center}
    \begin{tabular}{cccc}
      \includegraphics[width=0.24\hsize]{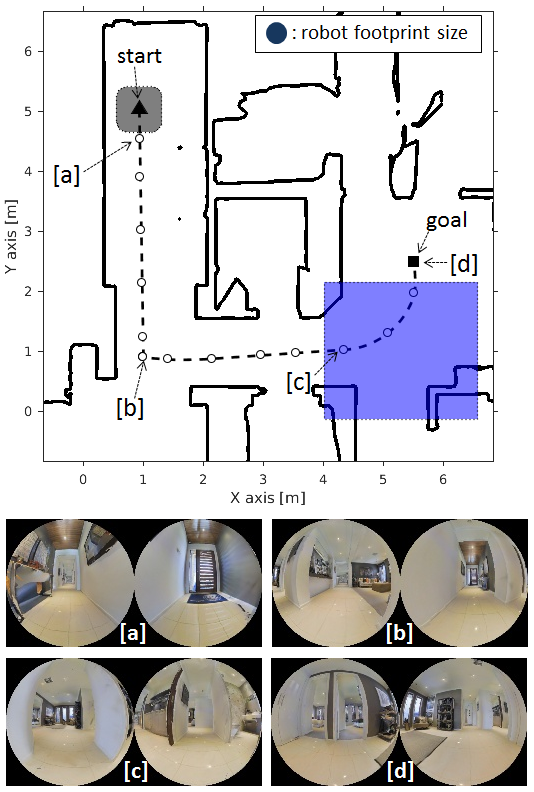} & \includegraphics[width=0.24\hsize]{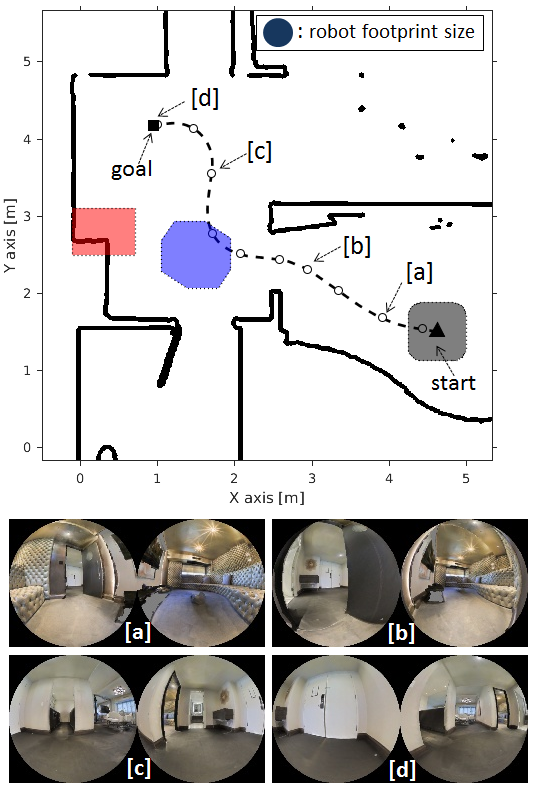} & \includegraphics[width=0.24\hsize]{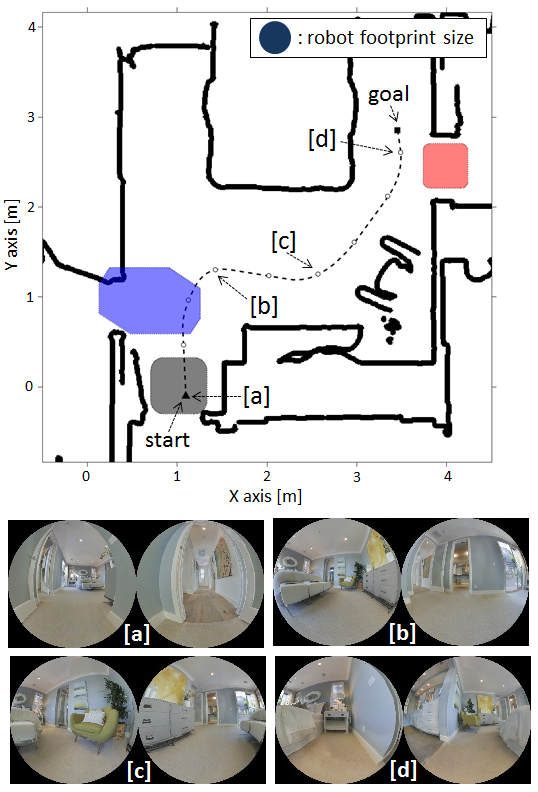} & \includegraphics[width=0.17\hsize]{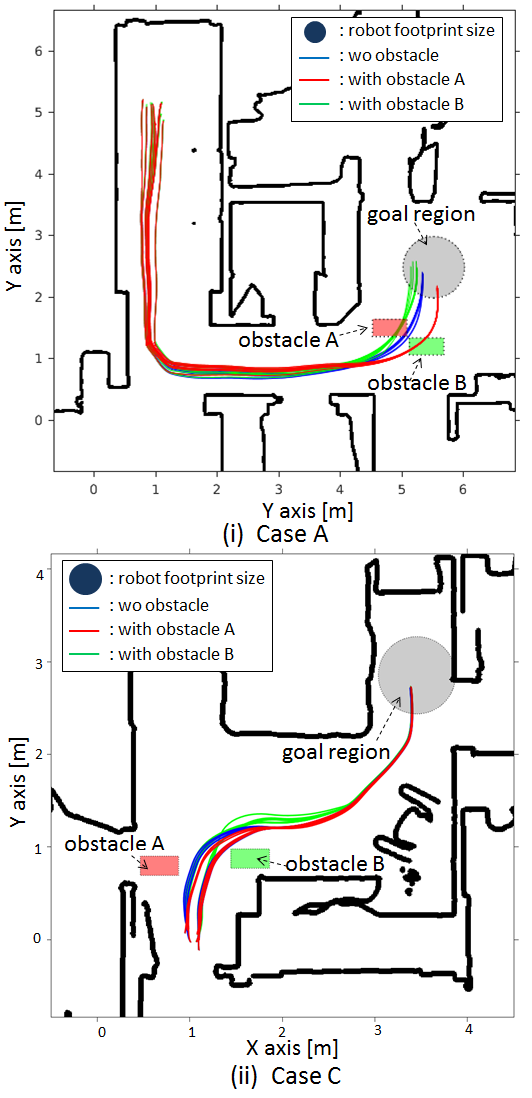} \\
      {\small $\left[ \mbox{a} \right]$ Case A {\small(corridor)} : 9.6 m} & {\small $\left[ \mbox{b} \right]$ Case B {\small(to dining room)} : 5.2 m} & {\small $\left[ \mbox{c} \right]$ Case C {\small(bed room)} : 5.5 m} & {\small $\left[ \mbox{d} \right]$ Robot trajectories}\\
    \end{tabular}
  \end{center}
\caption{\footnotesize Simulation environments for quantitative evaluation and robot trajectories of two examples. Top image in [a], [b] and [c] is the map of each environment. Bottom images show images from the visual trajectory, which is indicated in dashed line on the map.
[d] shows robot trajectories in (i) case A and (ii) case C in simulator with and without obstacle. The superimposed blue lines are the robot trajectories from 10 different initial poses without the obstacle. The red and green lines are the robot trajectories with the obstacle shown as the rectangular red and green region. The grey circle is the goal region, which is used to determine if the robot arrived at the goal.}
  \label{fig:traj}
  \end{figure*}
  
\subsection{Evaluation of PoliNet}

We evaluate PoliNet and compare it to various baselines, first as method to learn MPC-policies and then as core of our visual navigation approach. We now briefly introduce the baselines for this evaluation, while encouraging readers to refer to the original papers for more details.  

\textbf{MPC online optimization (back propagation)~\cite{wahlstrom2015pixels}:}
Our baseline backpropagates on the loss function $J$ to search for the optimized velocities instead of using a neural network. This approach is often used when the MPC objective is differentiable~\cite{wahlstrom2015pixels,hirose2014personal}. We evaluate three different number of iterations for the backpropagation, $n_{it}$ = 2, 20, and 100.

\textbf{Stochastic optimization~\cite{finn2017deep}:}
We use a cross-entropy method (CEM) stochastic optimization algorithm similar to~\cite{finn2017deep} as baseline of a different optimization approach. To do that, we sample $M$ sets of $N$ linear and angular velocities, and calculate $J$ for each velocity. Then, we select $K$ set of velocities with the smallest $J$ and resample a new set of $M$ from a multivariate Gaussian distribution of $K$ set of velocities. 
In our comparison to our approach we evaluate three sets of parameters of this method: ($M$, $K$, $n_{it}$) = (10, 5, 3), (20, 5, 5), and (100, 20, 5).

\textbf{Open loop control:} This baseline replays directly to the teleoperator's velocity commands in open loop as a trajectory. 

\textbf{Imitation Learning (IL)~\cite{codevilla2018end}:} We train two models to imitate the teleoperator's linear and angular velocity. The first model learns to imitate only the next velocity (comparable to PoliNet with $N=1$). The second model learns to imitate the next eight velocities (comparable to PoliNet with $N=8$).

\paragraph*{PoliNet as learned MPC}

Table~\ref{tab:sim_result} depicts the quantitative results of PoliNet as a learned MPC approach. Since our goal is to learn to emulate MPC, we report $J$, $J^{\textrm{\it img}}$, $J^{\textrm{\it trav}}$, and $J^{\textrm{\it ref}}$ of our approach and the baselines in 8000 cases randomly chosen from the test dataset.
In addition, we show the memory consumption and the inference time, crucial values to apply these methods in a real navigation setup. Note that the baseline methods except IL and open loop control employs same cost function of our method.

Both the backpropagation and the stochastic optimization baselines with largest number of iterations $n_{it}$ and bigger batch size $M$ can achieve the lowest cost values. However, these parameters do not compute at the \SI{3}{\hertz} required for real time navigation. To achieve that computation frequency we have to limit $M$ and $n_{it}$ to a small number leading to worse performance than PoliNet. 
Our method achieves a small cost value with less GPU memory occupancy (325 MB) and faster calculation speed (\SI{72.25}{\hertz}). Open loop control and imitation learning (IL) can also achieve small cost values with acceptable calculation speed and memory size but are less robust and reactive as we see in other experiments.



\paragraph*{PoliNet for Deep Visual Navigation}

To evaluate PoliNet as the core of our deep visual navigation approach, we perform three sets of experiments. In the first set we evaluate navigation in simulation with simulation-generated trajectories. In simulation we can obtain ground truth of the robot pose and evaluate accurately the performance. In the second set we evaluate in real world the execution of real world visual trajectories. In the third set we evaluate sim2real transfer, whether PoliNet-based navigation can reproduce in real world trajectories generated in simulation.

For the \textbf{first set} of experiment, \textbf{navigation in simulation}, we select three simulated environments (see Fig.~\ref{fig:traj}) and create simulated trajectories that generate visual trajectories by sampling images at \SI{0.333}{\hertz} (dashed lines). 
The robot start randomly from the gray area and needs to arrive at the goal. Obstacles are placed in blue areas (on trajectory) and red areas (off trajectory). To model imperfect control and floor slippage, we multiply the output velocities by a uniformly sampled value between $0.6$ and $1.0$. With this large noise execution we evaluate the robustness of the policies against noise.

Table~\ref{tab:PoliNet} depicts the success rate, the coverage rate (ratio of the arrival at each subgoal images), and SPL~\cite{anderson2018evaluation} of 100 trials for each of the three scenarios and the average. 
Here, the definition of the arrival is that the robot is in the range of $\pm$\SI{0.5}{\meter} of the position where the subgoal image was taken (note that the position of the subgoal image is only used for the evaluation). 

%
\begin{table}[h]
  \centering\
  \caption{Evaluations of PoliNet as learned MPC}
  \centering\
  \resizebox{1\columnwidth}{!}{    
  \label{tab:sim_result}
  \begin{tabular}{lr|c|c|c|c||c|c}\wcline{1-8}
       & & $J$ & $J^{\textrm{\it img}}$ & $J^{\textrm{\it trav}}$ & $J^{\textrm{\it ref}}$ & [MB] & [Hz]  \\ \wcline{1-8} 
       (a)Backprop.~\cite{wahlstrom2015pixels} & sim. & 0.436 & 0.284 & 0.240 & 0.322 & 1752 & 3.12 \\
       $(n_{it}=2)$ & real & 0.356 & 0.266 & 0.122 & 0.290 & & \\ \cline{1-8}
       (b)Backprop. & sim. & 0.270 & 0.230 & 0.038 & 0.210 & 1752 & 0.14 \\
       $(n_{it}=20)$ & real & 0.236 & 0.205 & 0.029 & 0.167 & & \\ \cline{1-8}
       (c)Backprop. & sim. & 0.220 & 0.199 & 0.017 & 0.122 & 1752 & 0.0275 \\
       $(n_{it}=100)$ & real & 0.205 & 0.185 & 0.016 & 0.119 & & \\ \cline{1-8}
       (d)Stochastic Opt.~\cite{finn2017deep} & sim. & 0.305 & 0.242 & 0.033 & 0.467 & 1728 & 3.66 \\
       $\small{(M=10, K=5, n_{it}=3)}$ & real & 0.279 & 0.226 & 0.031 & 0.380 & & \\ \cline{1-8}
       (e)Stochastic Opt. & sim. & 0.266 & 0.221 & 0.018 & 0.361 & 2487 & 1.59 \\
       $\small{(M=20, K=5, n_{it}=5)}$ & real & 0.247 & 0.208 & 0.018 & 0.307 & & \\ \cline{1-8}
       (f)Stochastic Opt. & sim. & 0.213 & 0.184 & 0.012 & 0.223 & 8637 & 0.39 \\
       $\small{(M=100, K=20, n_{it}=5)}$ & real & 0.205 & 0.181 & 0.011 & 0.183 & & \\ \cline{1-8}
       (g)Open loop control & sim. & $-$ & 0.220 & 0.087 & 0.000 & $-$ & $-$ \\
       & real & $-$ & 0.216 & 0.069 & 0.000 & & \\ \cline{1-8}       
       (h)Imitation learning & sim. & $-$ & 0.247 & 0.189 & 0.289 & 325 & 72.25 \\
       ($N$ = 8) & real & $-$ & 0.208 & 0.064 & 0.094 & & \\ \cline{1-8}
       (i)Our method, PoliNet & sim. & 0.277 & 0.221 & 0.062 & 0.245 & 325 & 72.25 \\
       & real & 0.214 & 0.180 & 0.035 & 0.161 & & \\ \cline{1-8}       
	\end{tabular}
	}
\end{table}

\begin{table*}[h]
  \centering
  \caption{Navigation with PoliNet and baselines in simulation (navigation success rate/subgoal coverage rate/SPL)}
  \resizebox{1.8\columnwidth}{!}{   
  \label{tab:PoliNet}
  \begin{tabular}{lr|c|c|c|c}\wcline{1-6}
       & & average & Case A : 9.6 m & Case B : 5.2 m & Case C : 5.5 m \\ \wcline{1-6} 
       Backprop.~\cite{wahlstrom2015pixels} & wo ob. & 0.000\,/ \,0.211\,/ \,0.000 & 0.000\,/ \,0.175\,/ \,0.000 & 0.000\,/ \,0.264\,/ \,0.000 & 0.000\,/ \,0.193\,/ \,0.000 \\
       $(n_{it}=2)$ & w/ ob. & 0.000\,/ \,0.205\,/ \,0.000 & 0.000\,/ \,0.169\,/ \,0.000 & 0.000\,/ \,0.260\,/ \,0.000 & 0.000\,/ \,0.187\,/ \,0.000 \\ \cline{1-6}
       Stochastic Opt.~\cite{finn2017deep} & wo ob. & 0.000\,/ \,0.147\,/ \,0.000 & 0.000\,/ \,0.102\,/ \,0.000 & 0.000\,/ \,0.184\,/ \,0.000 & 0.000\,/ \,0.156\,/ \,0.000 \\
       $(M=10, K=5, n_{it}=3)$ & w/ ob. & 0.000\,/ \,0.150\,/ \,0.000 & 0.000\,/ \,0.104\,/ \,0.000 & 0.000\,/ \,0.181\,/ \,0.000 & 0.000\,/ \,0.164\,/ \,0.000 \\ \cline{1-6}
       Open Loop & wo ob. & 0.023\,/ \,0.366\,/ \,0.023 & 0.040\,/ \,0.206\,/ \,0.040 & 0.030\,/ \,0.394\,/ \,0.030 & 0.000\,/ \,0.497 \,/ \,0.000  \\
       & w/ ob. & 0.010\,/ \,0.287\,/ \,0.010 & 0.000\,/ \,0.353\,/ \,0.000 & 0.030\,/ \,0.274\,/ \,0.030 & 0.000\,/ \,0.234\,/ \,0.000  \\ \cline{1-6}    
       Imitation learning  & wo ob. & 0.320\,/ \,0.721\,/ \,0.320 & 0.000\,/ \,0.501\,/ \,0.000 & 0.960\,/ \,0.981\,/ \,0.960 & 0.000\,/ \,0.683 \,/ \,0.000 \\
       ($N$ = 1)& w/ ob. & 0.297\,/ \,0.659\,/ \,0.291 & 0.000\,/ \,0.501\,/ \,0.000 & 0.890\,/ \,0.944\,/ \,0.873 & 0.000\,/ \,0.534\,/ \,0.000 \\ \cline{1-6}
       Imitation learning & wo ob. & 0.103\,/ \,0.592\,/ \,0.101 & 0.110\,/ \,0.618\,/ \,0.110 & 0.120\,/ 0.531\,/ \,0.113 & 0.080\,/ \,0.627\,/ \,0.079 \\
       ($N$ = 8)& w/ ob. & 0.310\,/ \,0.689\,/ \,0.310 & 0.100\,/ \,0.619\,/ \,0.099 & 0.800\,/ \,0.922\,/ \,0.799 & 0.030\,/ \,0.528\,/ \,0.030 \\ \cline{1-6}
       Zero-shot visual imitation(ZVI)\cite{pathak2018zero} & wo ob. & 0.433\,/ \,0.688\,/ \,0.432 & 0.000\,/ \,0.373\,/ \,0.000 & \textbf{1.000}\,/ \textbf{1.000}\,/ \,\textbf{1.000} & 0.300\,/ \,0.691\,/ \,0.297 \\
       & w/ ob. & 0.307\,/ \,0.551\,/ \,0.305 & 0.000\,/ \,0.361\,/ \,0.000 & 0.710\,/ \,0.883\,/ \,0.708 & 0.210\,/ \,0.410\,/ \,0.207 \\ \cline{1-6}
       {\it Vis. memory for path following}*~\cite{kumar2018visual} & wo ob. & {\it 0.858}\,/ \,------\,/ \,{\it 0.740} & {\it 0.916}\,/ \,------\,/ \,{\it 0.776} & {\it 0.920}\,/ ------\,/ \,{\it 0.834} & {\it 0.738}\,/ \,------\,/ \,{\it 0.611} \\
       *no collision detection & w/ ob. & {\it 0.800}\,/ \,------\,/ \,{\it 0.688} & {\it 0.942}\,/ \,------\,/ \,{\it 0.800} & {\it 0.660}\,/ \,------\,/ \,{\it 0.593} & {\it 0.798}\,/ \,------\,/ \,{\it 0.671} \\ \cline{1-6}         
       Our method & wo ob. & \textbf{0.997}\,/ \,\textbf{0.999}\,/ \,\textbf{0.996} & \textbf{0.990} \,/ \,\textbf{0.996}\,/ \,\textbf{0.989} & \textbf{1.000}\,/ \,\textbf{1.000}\,/ \,\textbf{1.000} & \textbf{1.000}\,/ \,\textbf{1.000}\,/ \,\textbf{1.000} \\
       & w/ ob. & \textbf{0.850}\,/ \,\textbf{0.916}\,/ \,\textbf{0.850} & \textbf{0.900}\,/ \,\textbf{0.981}\,/ \,\textbf{0.899} & \textbf{0.910}\,/ \,\textbf{0.963}\,/ \,\textbf{0.910} & \textbf{0.740}\,/ \,\textbf{0.805}\,/ \,\textbf{0.740} \\ \cline{1-6}       
	\end{tabular}
	}
	\vspace{-1.5em}
\end{table*}

Our method achieves 0.997 success rate without obstacles and 0.850 with obstacles on average and outperform all baseline methods. In addition, SPL for our method is close to the success rate, which means that the robot can follow the subgoal images without large deviation.
{\colornewparts Interestingly, in case B, IL with $N=8$  achieves better performance with obstacle than without obstacle. This is because the appearance of the obstacle helps the navigation in the area where the method fails without obstacle.
Note that, we show the results of \cite{kumar2018visual} without collision detection, because their evaluation doesn't support detecting collisions. We also implemented \cite{pathak2018zero} ourselves and trained on our dataset.

To verify that the results are statistically significant, we further evaluated our method compared to the strongest baseline, IL($N=8$) and ZVI\cite{pathak2018zero}, in seven additional random environments, and each 100 additional random runs as shown in the appendix. Our method obtained a total average success rate, coverage rate, and SPL of (0.979, 0.985, 0.979) without obstacles, and (0.865, 0.914, 0.865) with obstacles. This performance is higher than IL: (0.276, 0.629, 0.274), and ZVI: (0.574, 0.727, 0.573) without obstacles, and IL: (0.313, 0.643, 0.312), and ZVI: (0.463, 0.638, 0.461) with obstacles. These experiments further support the effectiveness of our method.}

Figure \ref{fig:traj}~[d] shows the robot trajectories in two scenarios with and without obstacle. The blue lines are the robot trajectories of 10 trials from the difference initial pose without obstacle. The red and green lines are the trajectories of 10 trials with the obstacle A (red) and the obstacle B (green). The grey circle is the area of the goal. Our method can correctly deviate from visual trajectory to avoid the obstacle and arrive at the goal area in these case without collision.

In the \textbf{second set} of experiments we evaluate PoliNet in \textbf{real world navigation} with and without previously unseen obstacles. We record a trajectory with the robot and evaluate it different days and at different hours so that the environmental conditions change, e.g. different position of the furniture, dynamic objects like the pedestrians, and changes in the lighting conditions. 
We normalize $(I_t^f, I_t^b)$ to have the same mean and standard deviation as $(\mathbf{I}_j^f, \mathbf{I}_j^b)$. Table~\ref{tab:r2r} shows the success rate and the coverage rate with and without obstacles in the original path. Our method can achieve high success rate and coverage rate for all cases and outperforms the baseline of imitation learning with eight steps by a large margin. (Note: other baselines cannot be used in this real time setup).

\begin{figure}[t]
  \centering
    \includegraphics[width=0.99\linewidth]{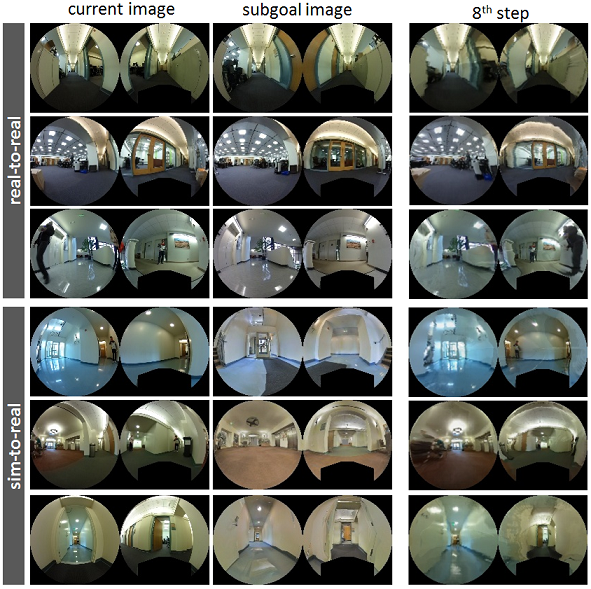}
	\caption{\footnotesize Visualization of the navigation performance of PoliNet in real word and simulation; \textit{Left}: current image from (real or simulated) robot; \textit{Middle}: next subgoal image; \textit{Right}: the VUNet-360 predicted image at 8-th step for velocities generated by PoliNet}
	\label{fig:img_polinet}
    \vspace{-1.5em}
\end{figure}

The first three rows of Fig.~\ref{fig:img_polinet} depict some exemplary images from navigation in real world. The figure depicts the current image (left), subgoal image (middle) and the predicted image at the eighth step by VUNet-360 conditioned by the velocities from PoliNet (right). 
There are some environment changes between the time the trajectory was recorded (visible in the subgoal image) and the testing time (visible in current image). For example, the door is opened in first example (top row), the light in one room is turned on and the brown box is placed at the left side in second example (second row), and the lighting conditions are different and a pedestrian is visible in the third example (third row). Even with these changes, our method generates accurate image predictions, close to the subgoal image. This indicates that PoliNet generates velocities that navigate correctly the robot towards the position where the subgoal image was acquired.


In the \textbf{third set} we evaluate \textbf{sim-to-real transfer}: using visual trajectories from the simulator to navigate in the corresponding real environment. 
We perform 10 trials for the trajectories at different days and times of the day.
The results of the sim-to-real evaluation are summarized in Table~\ref{tab:s2r}. The performance of our navigation method is worse than real-to-real, which is expected because there is domain gap between simulation and real world. Despite of that, our method can still arrive at the destination without collision in most of the experiments, indicating that the approach can be applied to generate virtual visual trajectories to be executed in real world.

The second three rows of Fig.~\ref{fig:img_polinet} depict some exemplary images from navigation in the sim-to-real setup. The discrepancies between the simulated images (subgoals) and the real images (current) are dramatic. For example, in 4th row, the black carpet is removed; in 5th and 6th row, there are big color differences. In addition, the door is opened in 6th row. To assess whether the velocities from PoliNet are correct, we compare the predicted 8th step image and subgoal image. Similar images indicate that the velocities from PoliNet allow to minimize visual discrepancy. Despite the changes in the environment, our deep visual navigation method based on PoliNet generates correctly velocities to minimize the visual differences.  

\begin{table}[h]
  \centering
  \caption{Navigation in real world}
  \centering
  \resizebox{1\columnwidth}{!}{
  \label{tab:r2r}
  \begin{tabular}{l|c|c|c|c}\wcline{1-5}
       & \multicolumn{2}{c|}{Case 1: 24.1 m} & \multicolumn{2}{c}{Case 2: 16.1 m} \\
       & wo ob. & w/ ob. & wo ob. & w/ ob.  \\ \cline{1-5}
       (h) IL ($N$=8) & 0.10 \,/ \,0.493 & 0.10 \,/ \,0.617 & 0.90 \,/ \,0.967 & 0.40 \,/ \,0.848 \\ \cline{1-5}
       (i) Our method & 1.00 \,/ \,1.000 & 0.80 \,/ \,0.907 & 1.00 \,/ \,1.000 & 0.80 \,/ \,0.910 \\ \cline{1-5}
	\end{tabular}
	}
\end{table}
\begin{table}[h]
  \centering
  \caption{Navigation in sim-to-real}
  \label{tab:s2r}
  \begin{tabular}{l|c|c|c}\wcline{1-4}
       & Case 3: 6.6 m & Case 4: 8.4 m & Case 5: 12.7 m \\ \wcline{1-4} 
       Our method & 0.90 \,/ \,0.983 & 0.80 \,/ \,0.867 & 0.80 \,/ \, 0.933 \\ \cline{1-4}
	\end{tabular}
\end{table}


%
%
%

\subsection{Ablation Study of Traversability Loss Generation}

In our method, $J^{\textrm{\it trav}}$ is one of the most important components for navigation with obstacle avoidance.
Table \ref{tab:ablation} is the ablation study for $J^{\textrm{\it trav}}$. We evaluate $J$, $J^{\textrm{\it img}}$, $J^{\textrm{\it trav}}$ and $J^{\textrm{\it ref}}$ for each weighting factor $q_{g}=0.0, 0.5, 1.0$ of $J^{\textrm{\it trav}}$ on the test data of both simulator's images and real images. In addition, we evaluate the model's navigation performance in simulator. 

We test our method 100 times in 3 different environments with and without obstacle. Average of the success rates (the ratio which the robot can arrive at the goal) are listed in the most right side.

Bigger $q_g$ can lead to smaller $J^{\textrm{\it trav}}$ and bigger $J^{\textrm{\it img}}$. However, the success rate of $q_g = 1.0$ is almost zero even for the environment without obstacle. Because too big $q_g$ leads the conservative policy in some cases. The learned policy tend to avoid narrow paths to keep $J^{\textrm{\it trav}}$ high, failing to arrive at target image. As a result, we decide to use the model with $q_g = 0.5$ for the all evaluations for our method.  

%
\begin{table}[h]
  \centering\
  \caption{Ablation Study of Traversability Loss Generation}
  \centering\
  \resizebox{1\columnwidth}{!}{    
  \label{tab:ablation}
  \begin{tabular}{lr|c|c|c|c||c|c}\wcline{1-8}
       & & $J$ & $J^{\textrm{\it img}}$ & $J^{\textrm{\it trav}}$ & $J^{\textrm{\it ref}}$ & with ob. & wo ob. \\ \wcline{1-8} 
       Our method & sim. & 0.244 & 0.217 & 0.161 & 0.261 & 0.810 & 0.780 \\
       $(q_{g} = 0.0)$ & real & 0.181 & 0.168 & 0.050 & 0.128 & & \\ \cline{1-8}       
       Our method & sim. & 0.277 & 0.221 & 0.062 & 0.245 & \textbf{0.997} & \textbf{0.850} \\
       $(q_{g} = 0.5)$ & real & 0.214 & 0.180 & 0.035 & 0.161 & \\ \cline{1-8}       
       Our method & sim. & 0.308 & 0.236 & 0.041 & 0.306 & 0.000 & 0.003 \\
       $(q_{g} = 1.0)$ & real & 0.246 & 0.194 & 0.032 & 0.196 & \\ \cline{1-8}         
	\end{tabular}
	}
\end{table}
%






%

\section{Conclusion and Future Work}
\label{sec:conclusion}

We presented a novel approach to learn MPC policies with deep neural networks and apply them to visual navigation with a 360\degree RGB camera. We presented PoliNet, a neural network trained with the same objectives as an MPC controller so that it learns to generate velocities that minimize the difference between the current robot's image and subgoal images in a visual trajectory, avoiding collisions and consuming less computational power than normal visual MPC approaches. Our experiments indicate that a visual navigation system based on PoliNet navigates robustly following visual trajectories not only in simulation but also in the real world. 

{\colornewparts One draw back of our method is that it fails occasionally to avoid large obstacles due to the following three main reasons: 1) In the control policy, traversability is a soft constraint balanced with convergence, 2) The predictive horizon is not enough to plan long detours and thus the agent cannot deviate largely from the visual trajectory, and 3) because large obstacles occupy most of the area of a subgoal image impeding convergence. In the future, we plan to experiment with traversability as hard constraint, increase the prediction horizon, and better transition between the subgoal images.}


%
%
%

\bibliographystyle{IEEEtranN}
\bibliography{IEEEabrv,references}

%
\newpage
\appendices
\section{Network Structure}
The details of the network structures are explained in the appendix.
\subsection{VUNet-360}
\begin{figure}[t]
  \centering
    \includegraphics[width=1.0\linewidth]{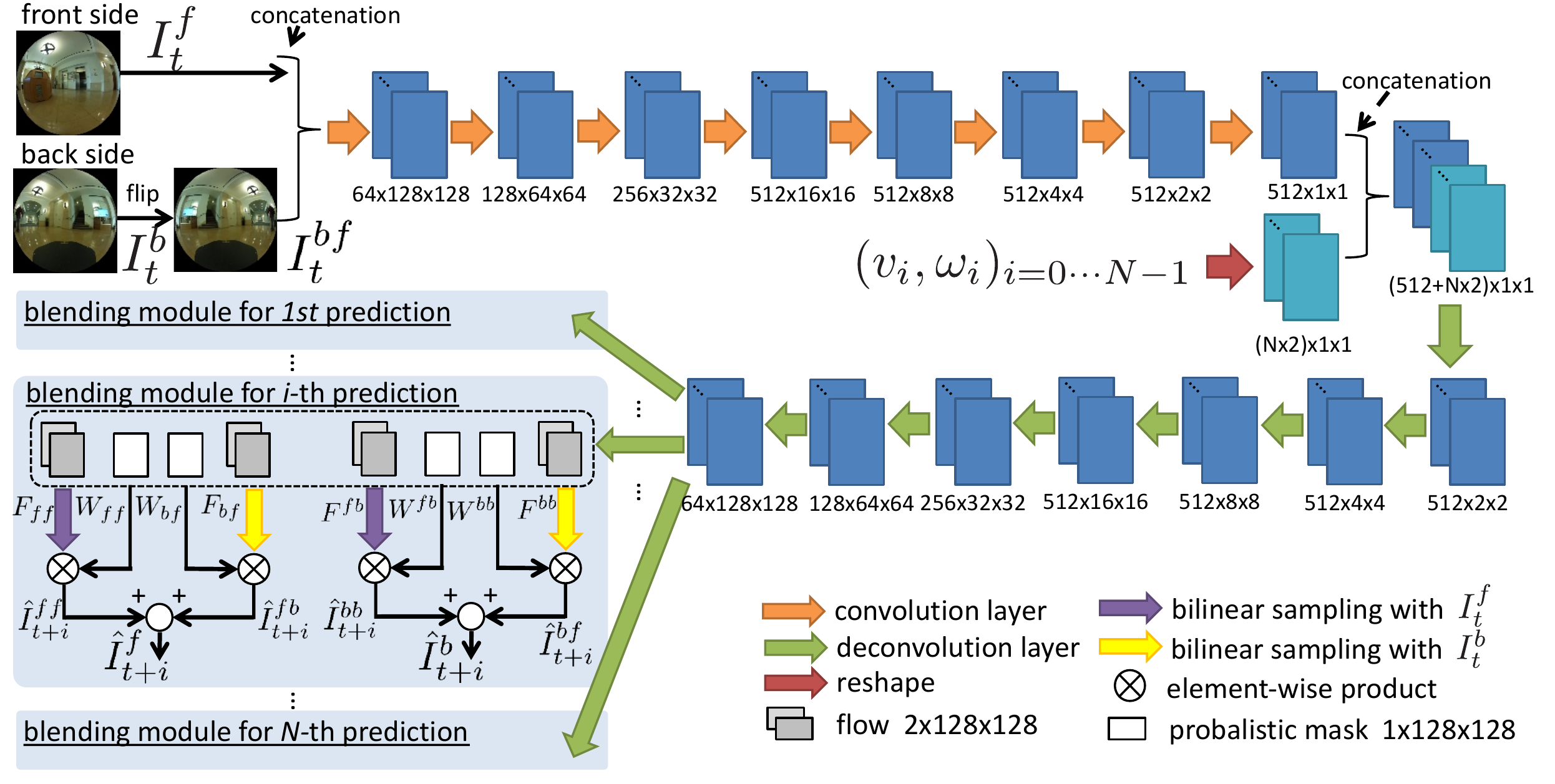}
	\caption{Network Structure of VUNet-360.}
	\label{fig:vunetx}
\end{figure}

VUNet-360 can predict $N$ steps future images by one encoder-decoder network, as shown in Fig.\ref{fig:vunetx}.
Different from the previous VUNet, VUNet-360 needs to have the network structure to merge the pixel values of the front and back side view for more precise prediction.
The concatenated front image $I_t^f$ and horizontally flipped back image $I_{t}^{bf}$ are fed into the encoder of VUNet-360.
The encoder is constructed by 8 convolutional layers with batch normalization and leaky relu function.
Extracted feature by the encoder is concatenated with $N$ steps virtual velocities $(v_i, \omega_i)_{i=0 \cdots N-1}$ to give the information about the robot pose in the future.
By giving 8 de-convolutional layers in the decoder part, $12N\times128\times128$ feature is calculated for $N$ steps prediction.
$12\times128\times128$ feature is fed into each blending module to predict the front and back side images by the synthesis of the internally predicted images\cite{zhou2016view}.
We explain the behavior of $i$-th blending module as the representative one.
For the prediction of the front image $\hat{I}_{t+i}^{f}$, we internally predict $\hat{I}_{t+i}^{ff}$ and $\hat{I}_{t+i}^{bf}$ by the bilinear sampling of $I_f^t$ and $I_b^t$ using two $2\times128\times128$ flows $F_{ff}$ and $F_{bf}$.
Then, we synthesize $\hat{I}_{f}^{t+i}$ by blending $\hat{I}_{ff}^{t+i}$ and $\hat{I}_{bf}^{t+i}$ as follows,
\begin{eqnarray}
\label{eq:vunet}
\hat{I}_{f}^{t+i} = \hat{I}_{ff}^{t+i} \otimes W_{ff} + \hat{I}_{bf}^{t+i} \otimes W_{bf},
\end{eqnarray}
where $W_{ff}$ and $W_{bf}$ are 1D probabilistic $1\times128\times128$ selection mask, and $\otimes$ is the element-wise product.
We use a softmax function for $W_{ff}$ and $W_{ff}$ to satisfy $W_{ff}(u,v) + W_{bf}(u,v) = 1$ for any same image coordinates $(u, v)$.
$\hat{I}_{t+i}^{b}$ can be predicted by same manner as $\hat{I}_{t+i}^{f}$, although the explanation is omitted.
%
%
\subsection{PoliNet}
PoliNet can generate $N$ steps robot velocities $(v_i, \omega_i)_{i=0 \cdots N-1}$ for the navigation.
In order to generate the appropriate velocities, PoliNet needs to internally understand i) relative pose between current and target image view, and ii) the traversable probability at the future robot pose.
On the other hand, computationally light network is required for the online implementation.
Concatenated $I_t^f$, $I_t^{bf}$, $\mathbf{I}_j^f$, and $\mathbf{I}_j^{bf}$ is given to 8 convolutional layers with batch-normalization and leaky relu activation function except the last layer.
In the last layer, we split the feature into two $N \times 1$ vector.
Then we give tanh($\cdot$) and multiply $v_{max}$ and $\omega_{max}$ for each vector to limit the linear velocity, $v_i (i=0 \cdots N-1)$ within $\pm v_{max}$ and the angular velocity within $\pm \omega_{max}$.
\begin{figure}[t]
  \centering
    \includegraphics[width=0.95\linewidth]{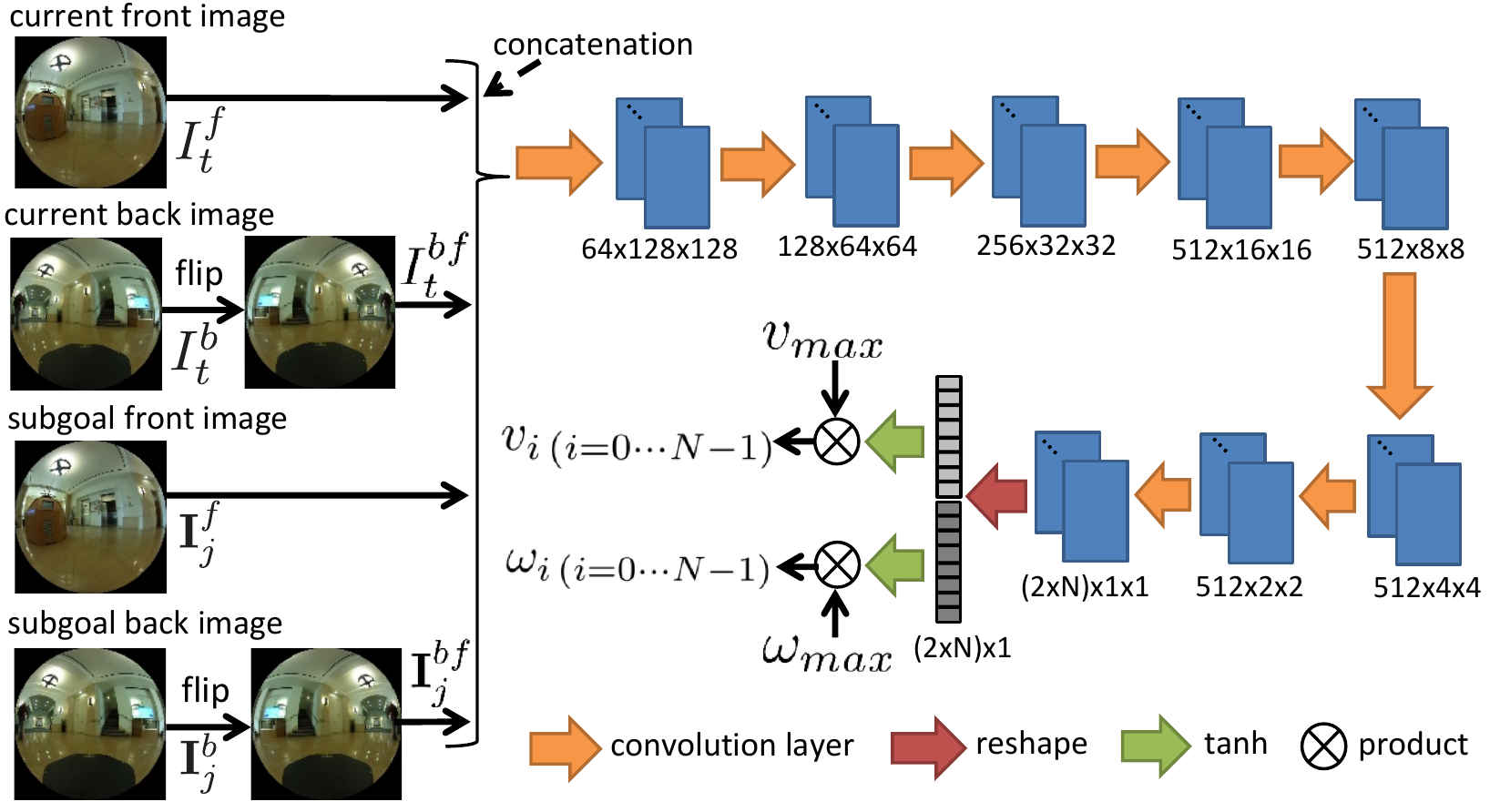}
	\caption{Network Structure of PoliNet.}
	\label{fig:polinet}
    \vspace{-1.5em}
\end{figure}
\section{Additional Evaluation for PoliNet}
\subsection{Additional Qualitative Result}
\begin{figure*}[t]
  \centering
    \includegraphics[width=0.99\linewidth]{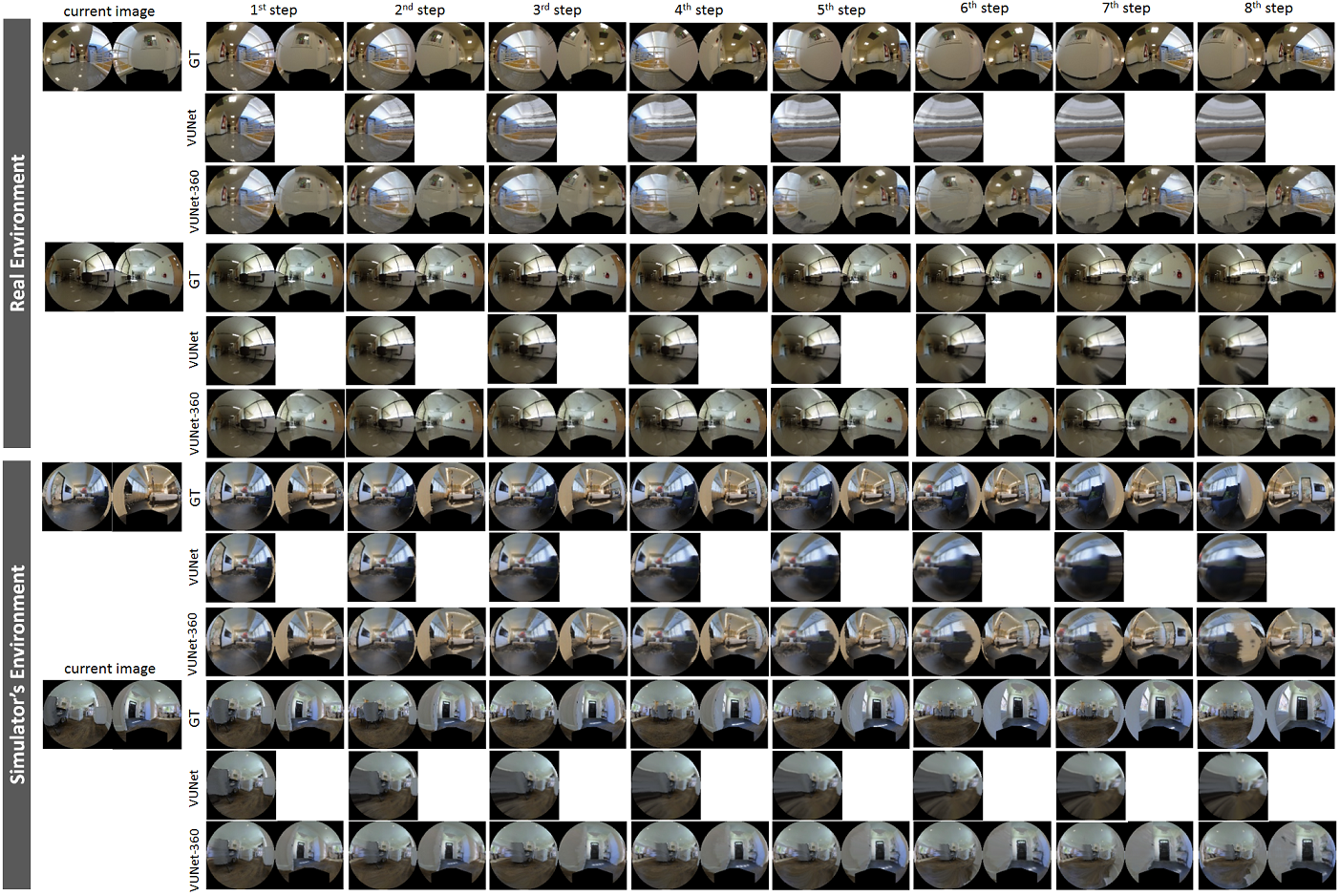}
	\caption{Predicted image by VUNet-360. Most left column shows current image. The other column shows the ground truth(GT), the predicted image by VUNet, and the predict image by VUNet-360 for each step. Top 6 rows are for test data of the real environment. Bottom 6 rows are for test data of the simulator's environment.}
	\label{fig:addvunet}
    \vspace{-1.5em}
\end{figure*}

Figure \ref{fig:addvunet} shows the additional 4 examples to qualitatively evaluate the performance of VUNet-360.
We feed the current image in most left side and tele-operator's velocities into the network to predict the images.
VUNet-360 can predict better images, which is close to the ground truth, even for 8-th step, the furthest future view.
On the other hand, VUNet can not correctly predict the environment, which is behind the front camera view.

\begin{figure*}[t]
  \centering
    \includegraphics[width=0.99\linewidth]{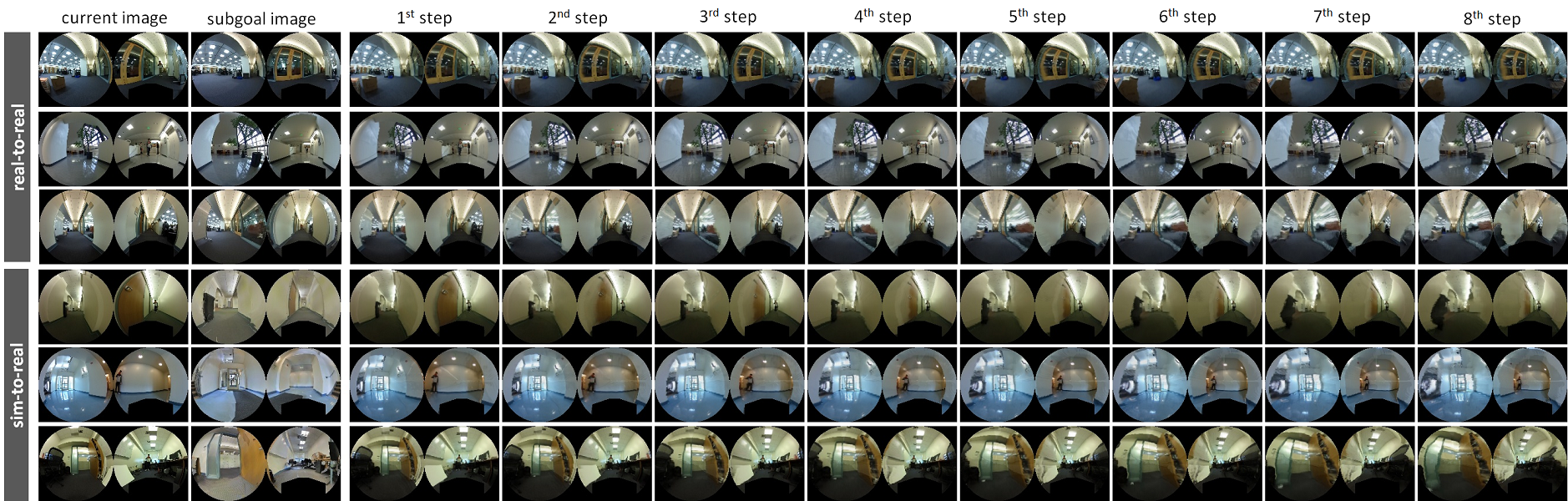}
	\caption{Visualization of the navigation performance of PoliNet in real environmet and simulation. First and second column show current and subgoal image. The other column shows the predicted images by VUNet-360 with the virtual velocities from PoliNet. Top 3 rows are for real-to-real. Bottom 3 rows are for sim-to-real.}
	\label{fig:addpolinet}
    \vspace{-1.5em}
\end{figure*}

In addition to Fig.\ref{fig:img_polinet}, Fig. \ref{fig:addpolinet} shows the additional 6 examples to evaluate PoliNet.
We show all predicted images for 8 steps to understand the exact behavior of PoliNet in Fig. \ref{fig:addpolinet}.
The time consecutive predicted images basically show continuous transition from the current image to subgoal image without untraversable situation.
It corresponds that the velocities by PoliNet can achieve the navigation from the current robot pose to the location of the subgoal image without collision and discontinuous motion.

\subsection{Additional Quantitative Result}
To verify that the results are statistically significant, we further evaluated our method compared to the strongest baseline, IL($N=8$) and ZVI\cite{pathak2018zero}, in seven additional random environments, and each 100 additional random runs.
The results of the seven new environments and the three previous environments in Table~\ref{tab:PoliNet} are shown in Table~\ref{tab:add}.
As shown in Table~\ref{tab:add}, our proposed method can work well in all environments.
\begin{table*}[h]
  \centering
  \caption{Additional evaluations in simulation}
  \centering
  \label{tab:add}
  \begin{tabular}{lc|c|c|c}\wcline{1-5}
       & & IL ($N$=8) & Zero-shot Vis. Imitation\cite{pathak2018zero} & Our method \\ \cline{1-5}
       Case A: 9.6 m & wo ob. & 0.110 \,/ \,0.618 \,/ \,0.110 & 0.000 \,/ \,0.373 \,/ \,0.000 & 0.990 \,/ \,0.996 \,/ \,0.989   \\ 
        & w/ ob. & 0.100 \,/ \,0.619 \,/ \,0.099 & 0.000 \,/ \,0.361 \,/ \,0.000 & 0.900 \,/ \,0.981 \,/ \,0.899  \\ \cline{1-5}
       Case B: 5.2 m & wo ob. & 0.120 \,/ \,0.531 \,/ \,0.113 & 1.000 \,/ \,1.000 \,/ \,1.000 & 1.000 \,/ \,1.000 \,/ \,1.000   \\ 
        & w/ ob. & 0.800 \,/ \,0.922 \,/ \,0.799 & 0.710 \,/ \,0.883 \,/ \,0.708 & 0.910 \,/ \,0.963 \,/ \,0.910  \\ \cline{1-5}
       Case C: 5.5 m & wo ob. & 0.080 \,/ \,0.627 \,/ \,0.079 & 0.300 \,/ \,0.691 \,/ \,0.297 & 1.000 \,/ \,1.000 \,/ \,1.000   \\ 
        & w/ ob. & 0.030 \,/ \,0.528 \,/ \,0.030 & 0.210 \,/ \,0.410 \,/ \,0.207 & 0.740 \,/ \,0.805 \,/ \,0.740  \\ \cline{1-5}
       Case D: 3.0 m & wo ob. & 0.790 \,/ \,0.847 \,/ \,0.774 & 0.690 \,/ \,0.803 \,/ \,0.684 & 0.970 \,/ \,0.973 \,/ \,0.970   \\ 
        & w/ ob. & 0.660 \,/ \,0.773 \,/ \,0.656 & 0.500 \,/ \,0.682 \,/ \,0.493 & 0.860 \,/ \,0.907 \,/ \,0.860  \\ \cline{1-5}
       Case E: 8.6 m & wo ob. & 0.700 \,/ \,0.782 \,/ \,0.700 & 0.990 \,/ \,0.992 \,/ \,0.990 & 0.950 \,/ \,0.965 \,/ \,0.950   \\ 
        & w/ ob. & 0.710 \,/ \,0.781 \,/ \,0.710 & 0.810 \,/ \,0.858 \,/ \,0.810 & 0.770 \,/ \,0.828 \,/ \,0.770  \\ \cline{1-5}
       Case F: 5.5 m & wo ob. & 0.770 \,/ \,0.847 \,/ \,0.770 & 1.000 \,/ \,1.000 \,/ \,1.000 & 1.000 \,/ \,1.000 \,/ \,1.000   \\ 
        & w/ ob. & 0.610 \,/ \,0.740 \,/ \,0.610 & 0.900 \,/ \,0.945 \,/ \,0.900 & 0.960 \,/ \,0.971 \,/ \,0.960  \\ \cline{1-5}
       Case G: 7.3 m & wo ob. & 0.000 \,/ \,0.300 \,/ \,0.000 & 0.610 \,/ \,0.715 \,/ \,0.610 & 0.940 \,/ \,0.953 \,/ \,0.940   \\ 
        & w/ ob. & 0.020 \,/ \,0.328 \,/ \,0.019 & 0.490 \,/ \,0.625 \,/ \,0.489 & 0.900 \,/ \,0.936 \,/ \,0.900  \\ \cline{1-5}
       Case H: 6.0 m & wo ob. & 0.190 \,/ \,0.466 \,/ \,0.189 & 0.150 \,/ \,0.399 \,/ \,0.145 & 0.940 \,/ \,0.963 \,/ \,0.940   \\ 
        & w/ ob. & 0.180 \,/ \,0.468 \,/ \,0.180 & 0.120 \,/ \,0.374 \,/ \,0.116 & 0.920 \,/ \,0.950 \,/ \,0.920  \\ \cline{1-5}
       Case I: 11.9 m & wo ob. & 0.000 \,/ \,0.495 \,/ \,0.000 & 0.000 \,/ \,0.293 \,/ \,0.000 & 0.990 \,/ \,0.996 \,/ \,0.990   \\ 
        & w/ ob. & 0.000 \,/ \,0.487 \,/ \,0.000 & 0.000 \,/ \,0.329 \,/ \,0.000 & 0.770 \,/ \,0.876 \,/ \,0.770  \\ \cline{1-5}
       Case J: 7.0 m & wo ob. & 0.000 \,/ \,0.778 \,/ \,0.000 & 1.000 \,/ \,1.000 \,/ \,1.000 & 1.000 \,/ \,1.000 \,/ \,1.000   \\ 
        & w/ ob. & 0.020 \,/ \,0.782 \,/ \,0.020 & 0.890 \,/ \,0.912 \,/ \,0.890 & 0.920 \,/ \,0.920 \,/ \,0.920  \\ \cline{1-5}
       Average & wo ob. & 0.276 \,/ \,0.629 \,/ \,0.274 & 0.574 \,/ \,0.727 \,/ \,0.573 & 0.979 \,/ \,0.985 \,/ \,0.979   \\ 
        & w/ ob. & 0.313 \,/ \,0.643 \,/ \,0.312 & 0.463 \,/ \,0.638 \,/ \,0.461 & 0.865 \,/ \,0.914 \,/ \,0.865  \\ \cline{1-5}         
	\end{tabular}
\end{table*}

\section{Additional Evaluation for VUNet-360}
In the main section, we only evaluate original VUNet and our predictive model VUNet-360, which can predict the multiple images in parallel.
In addition to them, we show the result of VUNet-360 for serial process, which can only predict the next image from previous image and virtual velocities.
To predict the far future image, we sequentially calculate the predictive model multiple times.
To train the predictive model for serial process, we evaluate two objectives.

The first objective is defined as
\begin{eqnarray}
J^{\textrm{\it VUNet}} = \frac{1}{2N_{\textrm{\it pix}}}(|I^f_{t+1} - \hat{I}^f_{t+1}| + |I^b_{t+1} - \hat{I}^b_{t+1}|),
\end{eqnarray}
where $(I^f_{t+1}, I^b_{t+1})$ are the ground truth future images and $(\hat{I}^f_{t+1}, \hat{I}^b_{t+1})$ are the VUNet predictions at $t+1$-th step.
In this first case, we only evaluate the difference between ground image and predicted image at next step.

The second objective is defined as
\begin{eqnarray}
J^{\textrm{\it VUNet}} = \frac{1}{2N\cdot N_{\textrm{\it pix}}}\sum_{i=0}^N (|I^f_{t+i} - \hat{I}^f_{t+i}| + |I^b_{t+i} - \hat{I}^b_{t+i}|)
\end{eqnarray}
where $(I^f_{t+i}, I^b_{t+i})_{i=1 \cdots N}$ are the ground truth future images and $(\hat{I}^f_{t+i}, \hat{I}^b_{t+i})_{i=1 \cdots N}$ are the VUNet predictions for $N$ steps.
In the second case, we evaluate $N(=8)$ steps to train the model. This is same objective as our method with parallel process.

\begin{figure*}[t]
  \centering
    \includegraphics[width=0.95\linewidth]{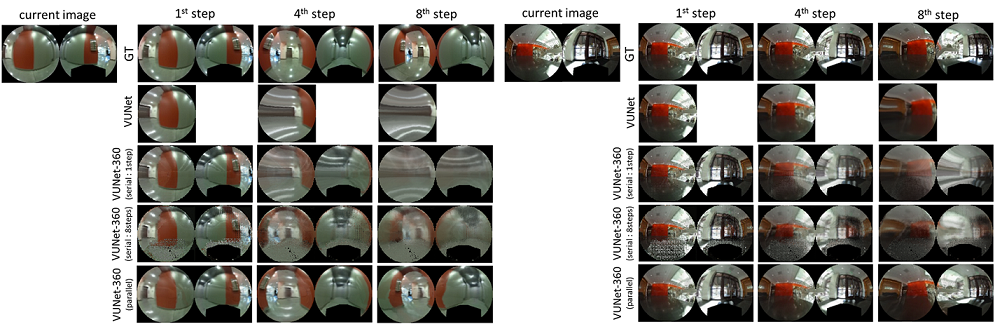}
	\caption{Predicted images by original VUNet\cite{hirose2019vunet}, VUNet-360 for serial process and VUNet-360 for parallel process(our method).}
	\label{fig:add_vunet}
    \vspace{-1.5em}
\end{figure*}

As can be observed in the Table our predictive model for parallel prediction overcomes all other methods with sequential process on L1 and SSIM. In addition, we can confirm that the parallel process predicts higher quality images as shown in Fig.\ref{fig:add_vunet}.

The model trained with the first objective can not correctly use the other side image (back side for front image prediction and front side for back image prediction), although the network has same blending module as VUNet-360 for parallel process.
The difference between $(I^f_{t+1}, I^b_{t+1})$ and $(\hat{I}^f_{t+1}, \hat{I}^b_{t+1})$ is too small to learn to use the other side image. Hence, the trained model tries to extend the edge of the image for the prediction instead of blending front and back images.

On the other hand, the model trained with the second objective seems to use the other side image to minimize the L1 loss. However, the predicted images present lower quality than our parallel processing model. We hypothesize the following causes for the blur:
\begin{enumerate}
 \item The parallel model can directly use the pixel information of the current raw image for all predicted images by bilinear sampling. However, the sequential process uses the pixels of the predicted image instead of the raw image, accumulating errors and possibly reducing the amount of information at each step of the process.
 \item The trained model for sequential process needs to receive not only raw image but also predicted images. We think that it is difficult for our network structure to encode both of them. On the other hand, our model for parallel process only receives the raw image. 
 \item The sequential process requires to calculate $N$ times layers for forward process and apply backpropagation for $N$ times deeper layers in sequential process. Hence, the sequential process is more difficult to train the model than the parallel one due to a more acute vanishing gradient problem.
\end{enumerate}
Because of the above reasons and faster calculation, we decided to use our proposed model with parallel prediction.

\begin{table*}[h]
  \centering
  \caption{Evaluation of predictive models.}
  \centering
  \resizebox{1.8\columnwidth}{!}{
  \label{tab:vunet_add}
  \begin{tabular}{lc|c|c|c|c|c|c|c|c|c}\wcline{1-11}
       & & avarage & 1st & 2nd & 3rd & 4th & 5th & 6th & 7th & 8th \\ \cline{1-11}
       VUNet & L1 & 0.132 & 0.058 & 0.085 & 0.110 & 0.131 & 0.149 & 0.163 & 0.176 & 0.186 \\ 
        & SSIM & 0.555 & 0.779 & 0.682 & 0.600 & 0.540 & 0.499 & 0.469 & 0.446 & 0.428 \\ \wcline{1-11}
       VUNet-360 & L1 & 0.121 & 0.060 & 0.087 & 0.107 & 0.123 & 0.135 & 0.145 & 0.153 & 0.160 \\ 
       (serial, 1 step) & SSIM & 0.581 & 0.783 & 0.683 & 0.613 & 0.565 & 0.532 & 0.508 & 0.491 & 0.477 \\ \wcline{1-11}
       VUNet-360 & L1 & 0.113 & 0.096 & 0.096 & 0.105 & 0.112 & 0.117 & 0.122 & 0.127 & 0.131 \\ 
       (serial, 8 steps) & SSIM & 0.543 & 0.611 & 0.588 & 0.563 & 0.544 & 0.528 & 0.515 & 0.504 & 0.494 \\ \wcline{1-11}
       {\bf VUNet-360 (our method)} & L1 & {\bf 0.088} & {\bf 0.057} & {\bf 0.068} & {\bf 0.078} & {\bf 0.086} & {\bf 0.093} & {\bf 0.100} & {\bf 0.106} & {\bf 0.114} \\ 
       {\bf (parallel, 8 steps)} & SSIM & {\bf 0.676} & {\bf 0.796} & {\bf 0.745} & {\bf 0.706} & {\bf 0.676} & {\bf 0.652} & {\bf 0.631} & {\bf 0.611} & {\bf 0.587} \\ \wcline{1-11} 
	\end{tabular}
	}
\end{table*}

\section{Analysis for results in Table II}
Although the baseline methods work well in each papers, the baseline methods often make mistakes in simulation of Table II. As the results, there are big gap between our proposed method and the baseline methods. In appendix, we explain the reasons for failures of each baselines.
\begin{description}
 \item[Backpropagation($n_{it}=2$)] We think that there are three main reasons for the very low performance:
 \begin{enumerate}
   \item In order to have a fair comparison we impose the same computational constraints on all algorithms: that they can be executed timely on real robot at, at least, 3 Hz. Conditioned on this constraint, the number of iterations in the backpropagation baseline is not enough to find a good velocity.
   \item The backpropagation method cannot use the  reference loss. 
   Therefore, the generated velocities are occasionally non-smooth and non-realistic and motivates some of the failures.
   \item As shown in Table I, first row, the computation time of the backpropagation method is enough to run at 3 Hz. However, this time introduces a delay between the moment the sensor signals are obtained and the velocity commands are sent that deteriorates the control performance. 
\end{enumerate}
 \item[Stochastic Optimization] The causes of this low performance are the same as for the previous baseline. Additionally, the predictive and control horizon of \cite{finn2017deep} is only three steps into the future, which is smaller than our predictive horizon of eight steps. This number of steps into the future is sufficient for the manipulation task of the original paper but seems too short for the navigation with obstacle avoidance.
 \item[Open Loop] In the 100 trials, we randomly change the initial robot pose and slip ratio. These uncertainties definitely deteriorate the performance of open loop control.
 \item[Imitation learning ($N$=1, and $N$=8)] It is known that imitation learning and behavior cloning often accumulate errors that can lead to finally fail in the task. We can find a few cases, which the imitation learning can achieve the high success rate.  However, the average is quite lower than our method. We think that our task is one of the inappropriate tasks for the imitation learning.
 \item[Zero-shot visual imitation(ZVI)] ZVI often collides both with and without obstacle because, different to ours, it does not penalize explicitly non-traversable area and because the prediction in the forward consistency loss comes always from the raw image at previous step, which may not be enough to deviate to avoid the obstacles.
 \item[Visual memory for path following] Although their performance is better than the other baselines, their original code doesn't include the collision detection module for the evaluation. It means that we didn't stop the evaluation even if their method collides with the obstacles. And, it seems that their control input is too discrete to realize the navigation in narrow space.
 
\end{description}
%

%
%

\end{document}